\documentclass[lettersize,journal]{IEEEtran}
\usepackage{amsmath,amsfonts}
\pdfoutput=1
\usepackage{algorithmic}
\usepackage{algorithm}
\usepackage{array}
\usepackage[caption=false,font=normalsize,labelfont=sf,textfont=sf]{subfig}
\usepackage{textcomp}
\usepackage{stfloats}
\usepackage{url}
\usepackage{verbatim}
\usepackage{graphicx}
\usepackage{cite}
\usepackage{amsmath}
\usepackage{multirow}
\usepackage{booktabs}
\usepackage{bm}
\usepackage{amssymb}
\usepackage{pifont}
\usepackage[dvipsnames]{xcolor}
\usepackage{colortbl}
\newcommand{\cmark}{\ding{51}}%
\newcommand{\xmark}{\ding{55}}%
\definecolor{mygray}{gray}{.9}
\definecolor{palegreen}{HTML}{F6FAF3}
\definecolor{tabhighlight}{HTML}{e5e5e5}

\usepackage[utf8]{inputenc} % allow utf-8 input
\usepackage[T1]{fontenc}    % use 8-bit T1 fonts
\usepackage{url}            % simple URL typesetting
\usepackage{amsfonts}       % blackboard math symbols
\usepackage{nicefrac}       % compact symbols for 1/2, etc.
\usepackage{microtype}      % microtypography

\usepackage{graphicx}
\usepackage{enumitem}
\usepackage{multicol}
\usepackage{adjustbox}
\usepackage{diagbox}
\usepackage[hidelinks]{hyperref}

\begin{document}

\title{Multi-modal Mutual-Guidance Conditional Prompt Learning \\ for Vision-Language Models}

\author{Shijun Yang, Xiang Zhang, Wanqing Zhao, Hangzai Luo, Sheng Zhong, Jinye Peng, Jianping Fan% <-this % stops a space
\thanks{This research was supported by the National Natural Science Foundation of China under Grant 62306237 and in part by the 2024 Graduate Research Innovation Project of Northwest University of China under Grant CX2024220.}

\thanks{Shijun Yang, Xiang Zhang, Wanqing Zhao, Hangzai Luo, Sheng Zhong, Jinye Peng, and Jianping Fan are with the School of Information and Technology, Northwest University, Xi’an, Shaanxi 710127, China (e-mail: yangshijun@stumail.nwu.edu.cn; xiangz@nwu.edu.cn; zhaowq@nwu.edu.cn; hzluo@nwu.edu.cn; szhong@nwu.edu.cn; pjy@nwu.edu.cn; jfan@nwu.edu.cn).}

\thanks{Corresponding author: Xiang Zhang. Email: xiangz@nwu.edu.cn.}
 
 % <-this % stops a space
}

\markboth{Journal of \LaTeX\ Class Files,~Vol.~14, No.~8, August~2021}%
{Shell \MakeLowercase{\textit{et al.}}: A Sample Article Using IEEEtran.cls for IEEE Journals}



\maketitle

\begin{abstract}
Prompt learning facilitates the efficient adaptation of Vision-Language Models (VLMs) to various downstream tasks. However, it faces two significant challenges: (1) inadequate modeling of class embedding distributions for unseen instances, leading to suboptimal generalization on novel classes; (2) prevailing methodologies predominantly confine cross-modal alignment to the final output layer of vision and text encoders, which fundamentally limits their capacity to preserve topological consistency with pre-trained multi-modal embedding spaces. To this end, we introduce MuGCP (Multi-modal Mutual-Guidance Conditional Prompt Learning), a novel paradigm designed for conditional prompt generation. MuGCP leverages Multi-modal Large Language Models (MLLMs) as conditional prompt learners to adaptively generate Semantic Conditional Prompts (SCP) that incorporate rich, fine-grained high-level semantic knowledge for image instances. To ensure effective alignment and interaction across the multi-modal space of Vision-Language Models (VLMs), we introduce the Attention Mutual-Guidance (AMG) module, which facilitates interactions between visual and semantic information. Through mutual guidance, the AMG module generates Visual Conditional Prompts (VCP), enhancing the model's performance in multi-modal tasks. Additionally, we present a Multi-Prompt Fusion (MPF) mechanism that integrates SCP and VCP with contextual prompts, ensuring seamless coordination among the different prompts and enhancing the modeling of class embeddings and instance-specific knowledge. Our MuGCP outperforms existing state-of-the-art methods on 14 different datasets. The code will be made available after publication.
\end{abstract}


\section{Introduction}
Vision-Language Models (VLMs), such as CLIP~\cite{radford2021learningCLIP} and ALIGN~\cite{jia2021scaling}, are trained on large-scale image-text pairs, leveraging rich natural language supervision to perform reasoning over open-set visual concepts. These models typically use manually designed prompt templates, embedding the class name of the downstream task within the template, such as ``a photo of a $\left \{\textrm{class name}\right \}$''. The text encoder then generates a text embedding for the prompt, which is matched with the image embedding generated by the image encoder to make predictions. Although carefully crafted prompts can significantly improve the performance of VLMs, the process of manually designing these templates is time-consuming and labor-intensive, requiring extensive linguistic adjustments to ensure accurate alignment between the prompt and the image. 

\begin{figure}[t]
\centering
\includegraphics[width=\linewidth]{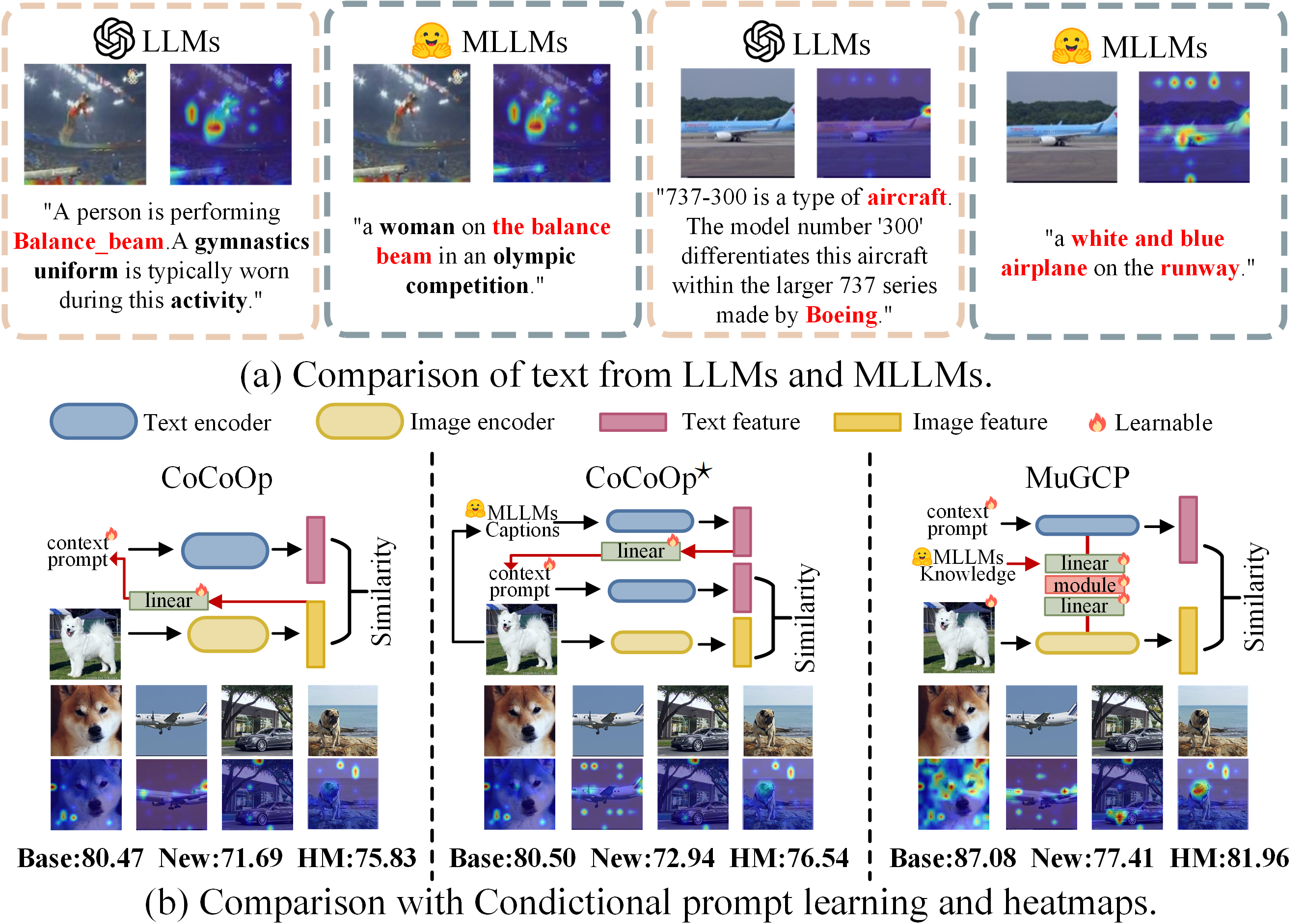} 
\caption{(a) Comparison of text from LLMs and MLLMs: LLMs generate class descriptions, and MLLMs produce instance-level descriptions, as indicated by heatmaps and word importance. (b) Comparison with existing conditional prompt learning frameworks and heatmaps: CoCoOp~\cite{zhou2022conditionalCoCoOp} (left) is an image conditional prompt learning. In $\text{CoCoOp}^{\star}$ (middle), $\star$ indicates replacing the image features of each instance with the text features of the image captions, achieving a 1.25\% improvement in generalization to new classes.}
\label{fig:Introduction1}
\end{figure}

Recently, methods such as prompt learning~\cite{zhou2022learningCoOp,zhou2022conditionalCoCoOp,zheng2023large,zang2024overcoming} and adapter learning~\cite{yu2023task,gao2024clip,li2024graphadapter} have introduced learnable weight vectors as substitutes for manual word tuning. These methods enable the effective adaptation of the input-output feature space of frozen VLMs encoder using few-shot data, avoiding catastrophic forgetting and overfitting~\cite{shu2023clipood,zhou2022learningCoOp} that may arise from fine-tuning entire VLMs, thereby improving generalization performance on downstream tasks. Among them, prompt learning has gained significant attention for its efficiency in adapting to and generalizing across diverse tasks. It can be roughly classified into two categories: (1) Domain-shared prompts. Domain-shared prompts involve learning contextual prompts that are shared across all categories or instances (images), irrespective of whether they belong to the text branch~\cite{zhou2022learningCoOp,yao2024tcp,zhang2024dept}, vision branch~\cite{zang2022unified,jia2022visual}, or both~\cite{khattak2023maple,wang2024bilateral}. However, these methods focus on a limited set of seen classes, failing to capture broader, image-specific information. Furthermore, by aligning vision and text modalities at the feature level of the final output layer, VLMs tend to deviate from their pre-trained multi-modal space. (2) Image-conditional prompts. Image-conditional prompts~\cite{zhou2022conditionalCoCoOp} integrate image features with text contextual prompts, providing instance-level knowledge that enhances prompt robustness and generalization to new classes. Despite using a bottleneck layer to bridge the image and text feature spaces, individual image features remain susceptible to single-sample bias and lack class embeddings distribution modeling. This limits the model’s ability to generalize to new (unseen) images or classes and increases the risk of overfitting.

To explicitly model the distributions of class embedding, recent advancements in Multi-modal Large Language Models (MLLMs)~\cite{li2023blip, zhu2023minigpt, liu2024visual} have enabled them to act as ``domain experts'' for individual image instances, providing corresponding image captions, as shown in Fig.~\ref{fig:Introduction1}(a), which include instance-specific information (e.g., ``olympic competition''), coarse-grained category information (e.g., ``balance beam''), and general knowledge (e.g., ``a woman''). Image captions not only describe the object information in the image but also capture high-level associative information, such as the semantic relationships and contextual dependencies between objects and their surroundings or other objects. For example, the caption ``a white and blue airplane on the runway'' reflects the spatial semantic relationship between the object ``airplane'' and the scene ``runway''. Compared to the image features used in CoCoOp~\cite{zhou2022conditionalCoCoOp}, this high-level semantic information in the image caption helps prompt learning to acquire complex semantic knowledge that goes beyond individual image feature, enhancing the understanding of object categories in scenes and contexts, thereby further optimizing the modeling of class embedding distributions in VLMs (as illustrated in Fig.~\ref{fig:Introduction1}(b)). It is worth noting that  MLLMs can only provide coarse-grained category information, as their pre-training data does not align with the task-specific data of downstream tasks. Even when a set of reference class names is provided, MLLMs often struggle to infer the correct fine-grained class labels accurately. Recent research~\cite{zhang2024visually} indicates that key classification information is encoded in the latent space of MLLMs, and effectively decoding these features requires large-scale fine-tuning on downstream training data. However, this fine-tuning process is computationally expensive and time-consuming.

To address the limitations of previous methods, we propose MuGCP, a novel multi-modal mutual-guidance conditional prompt learning framework. MuGCP enhances class embedding modeling by integrating Semantic Conditional Prompts (SCP) and Visual Conditional Prompts (VCP) while facilitating instance-level alignment and interaction within the multi-modal space through our proposed Attention Mutual-Guidance (AMG) module and Multi-Prompt Fusion (MPF) mechanism, thereby improving overall performance. Specifically, MuGCP leverages the few-shot learning capability of VLMs to adaptively decode the internal feature representations of MLLMs, thereby generating SCP rich in fine-grained high-level semantic knowledge for each instance. This addresses the challenge of explicitly modeling class embedding distributions in VLMs while avoiding the computational overhead of fine-tuning MLLMs on large-scale datasets. The AMG module generates semantically enriched VCP, which, together with SCP, dynamically aligns the representation spaces of text and vision, facilitating information fusion, feature filtering, and nonlinear modeling. The MPF mechanism integrates VCP and SCP with contextual prompts, leveraging the frozen transformer layers within the encoder to capture discriminative features and ensure effective coordination, improving learning and modeling between class embeddings and instance-specific knowledge. Finally, text augmentation based on MLLMs is used to enforce consistency constraints during prompt learning, further improving model robustness.

Our main contributions are summarized below as follows:
\begin{itemize}

\item We present MuGCP, an effective and flexible framework that leverages MLLMs as multi-modal conditional prompt learners to adaptively enhance the ability of prompt learning methods to explicitly model class embedding distributions. Additionally, MuGCP can seamlessly replace other mainstream MLLMs, facilitating their integration into VLMs.

\item The AMG module and MPF mechanism are designed to enhance instance-level cross-layer and cross-modal mutual-guidance interactions, realize information fusion, feature filtering, and nonlinear modeling, and combine them with contextual prompts to coordinate the modeling of class embeddings and instance-specific knowledge.

\item Experiments show that MuGCP outperforms existing state-of-the-art methods in both zero-shot generalization and few-shot classification across 14 different datasets.

\end{itemize}

\begin{figure*}[ht]
\centering
\includegraphics[width=0.95\textwidth]{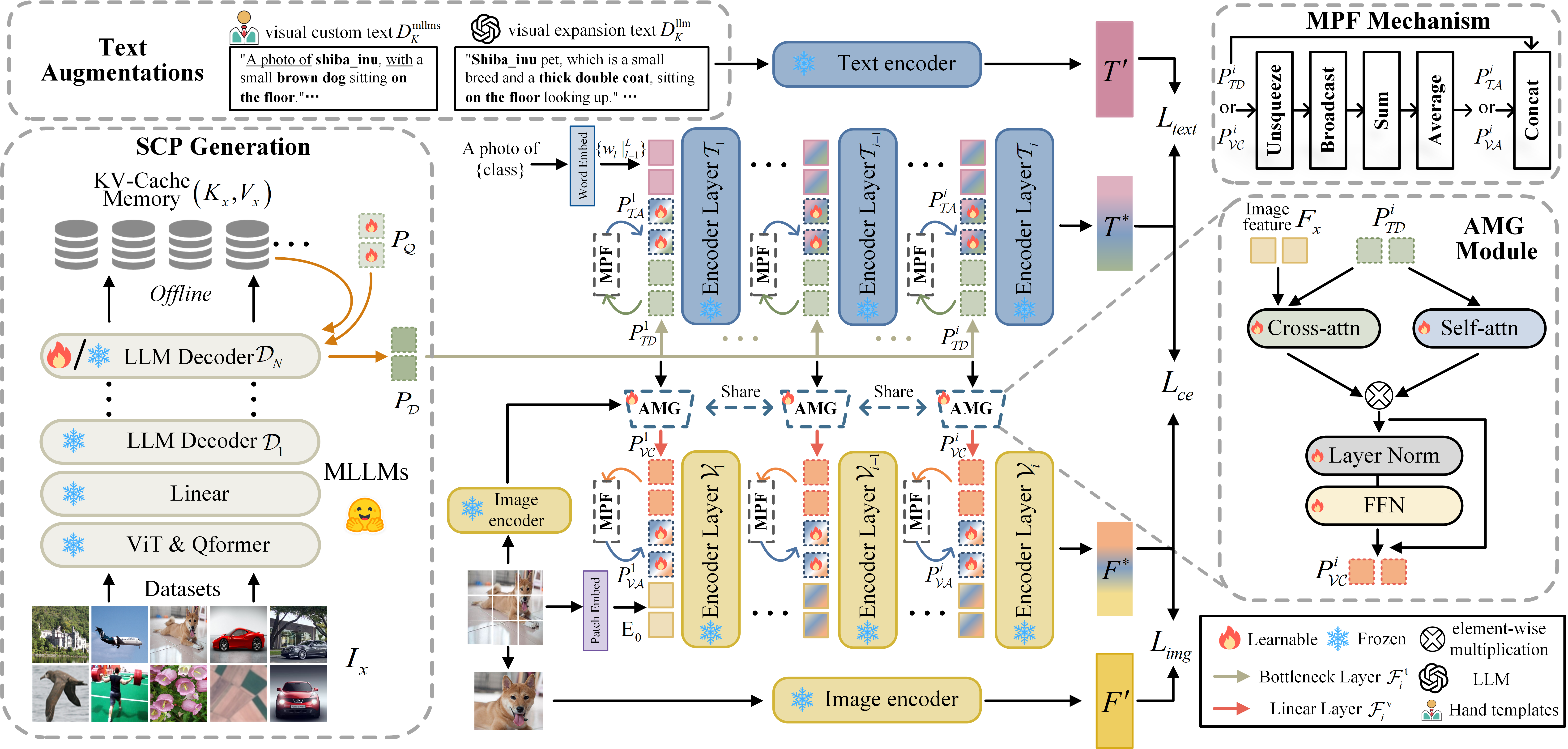}
\caption{The framework of Multi-modal Mutual-Guidance Conditional Prompt Learning (MuGCP). MuGCP leverages Multi-modal Large Language Models (MLLMs) as conditional prompt learners to generate \textbf{Semantic Conditional Prompts (SCP) $P_{\mathcal{TD}}^{i}$}. The \textbf{Attention Mutual Guidance (AMG) Module} generates \textbf{Visual Conditional Prompts (VCP) $P_{\mathcal{VC}}^{i}$} and facilitates prompt-level interactions between SCP and VCP, achieving information fusion, feature filtering, and nonlinear modeling. By adopting \textbf{Multi-Prompt Fusion (MPF) Mechanism}, MuGCP aligns prompts with word embeddings, patch embeddings, and instance knowledge, refining the learning and modeling of class embeddings and instance-specific knowledge. Finally, training with \textbf{Text Augmentation} and \textbf{Loss Function} constraints incrementally enhances the generalization capability of VLMs on downstream tasks through the proposed components.}
\label{OverView}
\end{figure*}

\section{Related Works}
\noindent\textbf{Vision-Language Models (VLMs).} VLMs have made significant progress in the field of visual representation learning by leveraging rich image-text pairs data. These models aim to learn rich visual representations that are guided by natural language, aligning text and image features within a shared embedding space. For instance, models such as CLIP~\cite{radford2021learningCLIP} and ALIGN~\cite{jia2021scaling} harness large-scale image-text datasets, demonstrating robust multi-modal knowledge and facilitating cross-domain transfer learning. This progress has enabled efficient solutions for a range of downstream tasks, including image classification~\cite{CLIP_GPT,li2024graphadapter,wang2024pedestrian}, text-based image generation~\cite{ramesh2021zero,rombach2022high}, and semantic segmentation~\cite{dong2023maskclip,yu2024convolutions,zhang2024language}. Despite their impressive generalization capabilities, effectively adapting pre-trained VLMs to specialized downstream tasks and further enhancing their performance remains a significant challenge.

\noindent\textbf{Prompt Learning.} Prompt learning has gained significant attention for its effectiveness in integrating methods such as adapters and its adaptability across different tasks. CoOp~\cite{zhou2022learningCoOp}, as a pioneering work, first introduced domain-shared contextual prompt learning to VLMs, enabling the transfer of task-specific knowledge to downstream tasks. Some works have improved domain-shared prompt learning from various aspects, including multi-modal prompts~\cite{khattak2023maple}, plug-and-play modules~\cite{ma2023understanding,zhang2024dept,zang2024overcoming,ren2024modality}, external knowledge enhancement~\cite{zhang2023prompt,zheng2023large}, and knowledge distillation~\cite{li2024promptkd}. On the other hand, image-conditional prompt learning~\cite{zhou2022conditionalCoCoOp} dynamically adjust contextual prompts based on the specific knowledge provided by image features, enhancing generalization to new classes. Despite these advancements, existing prompt learning methods often fail to provide useful semantic information for test domain instances, and most of them align the vision and text modalities at the feature level within the encoder's final output layer, causing VLMs to deviate from their pre-trained multi-modal space. In contrast, we utilize MLLMs to adaptively generate semantic conditional prompts rich in high-level semantic information, effectively modeling the class embedding distributions. Additionally, the attention mutual-guidance module enables semantic and visual information to mutually guide each other, dynamically adjusting and bridging multi-modal representation spaces. The multi-prompt fusion mechanism further ensures effective coordination between multiple prompts and embedding information.

\noindent\textbf{External Knowledge Enhanced VLMs.} Large Language Models (LLMs)~\cite{llama-v1,llama-v2,gpt-4} are frequently utilized as external knowledge bases for VLMs, providing class-specific knowledge, such as~\cite{CuPL,KAPT,zhang2023prompt,Saha_2024_CVPR} focuses on generating sentences relevant to class names and extracting critical information from text features to adapt to downstream tasks. Nevertheless, once each descriptive sentence is encoded by a text encoder into a sentence embedding, it becomes fixed and lacks flexibility. Moreover, these methods often rely on matching the average text embedding, calculated from all descriptive sentences for a given category, with the image embedding of the images to make predictions. This average text embedding is often inadequate for representing all instances within the category. Recently, Multi-modal Large Language Models (MLLMs)~\cite{chen2023minigpt,liu2024visual} have garnered significant attention for their ability to not only align vision and text modalities effectively but also to understand specific images and generate corresponding descriptive text. FineR~\cite{liu2024democratizing} employs MLLMs and LLMs iteratively to infer sub-category names, demonstrating the models' capability in classification tasks. TGP-T-F~\cite{tan2024compound} utilizes MiniGPT-4~\cite{zhu2023minigpt} to generate image descriptions, which are then used to optimize prompts, highlighting the role of these models in enhancing prompt engineering. Current research in this domain predominantly concentrates on acquiring class-level information to enhance the model's generalization for specific categories. This strategy, while valuable, often results in increased sensitivity of fine-tuned models to class shifts, adversely affecting their generalization performance. Instead, extracting external knowledge should be grounded in a comprehensive understanding of the individual image instance itself, along with the class information to which the instance belongs, which helps mitigate the effects of class shifts and provides a more nuanced representation of each instance, thereby enhancing the generalization capabilities of VLMs. To this end, we leverage MLLMs to dynamically capture high-level association information such as the semantic relationship and contextual dependency of each image instance in a few-shot learning manner. This reduces sensitivity to class shifts and enables a more holistic understanding of each instance. Subsequently, VLMs utilize conditional prompts provided by MLLMs, which are rich in fine-grained high-level semantic information, to assist in the generalization of downstream tasks.

\section{Methodolgy}
The overall framework of our MuGCP is shown in Fig.~\ref{OverView}. MuGCP leverages MLLMs as conditional prompt learners to adaptively produce Semantic Conditional Prompts (SCP), which are enriched with fine-grained high-level semantic knowledge for image instances. The generation of Visual Conditional Prompts (VCP) is facilitated by the Attention Mutual-Guidance (AMG) module, which enables cross-layer and cross-modal instance-level prompt interactions, aligning multi-modal feature spaces while generating VCP that are visually informed and tightly coupled with semantic context. Additionally, the Multi-Prompt Fusion (MPF) mechanism integrates SCP, VCP, and context prompts with word or patch embeddings, enhancing the learning and modeling of class embeddings and instance-specific knowledge, progressively enhancing the generalization ability of VLMs for downstream tasks. Finally, we introduce an augmented text generation strategy based on MLLMs, which effectively mitigates issues related to overfitting and multi-modal space drift, ensuring robust model performance.

\begin{figure}[t]
\centering
\includegraphics[width=\linewidth]{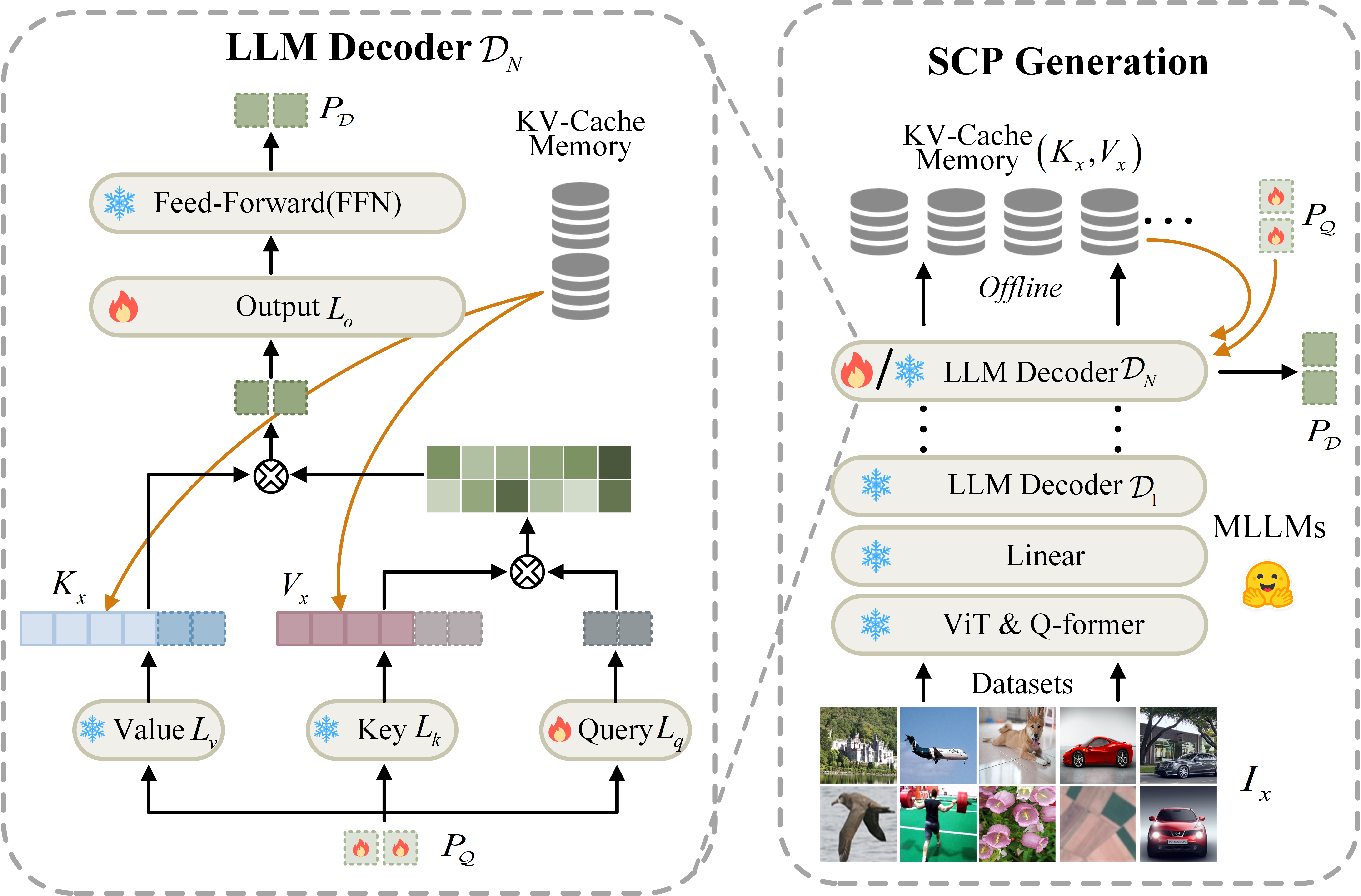} 
\caption{The Generation Process Diagram of Semantic Conditional Prompts.}
\label{fig:SCP}
\end{figure}

\subsection{Mutual-Guidance Conditional Prompt Learning}

\noindent\textbf{Semantic Conditional Prompts (SCP) Generation.} To provide conditional prompts with high-level semantic knowledge for different instances, image captions generated by MLLMs as ``domain experts'', which are rich in various types of information, seem like a promising approach. As shown in Fig.~\ref{fig:Introduction1}(b), $\text{CoCoOp}^{\star}$ encodes the image caption into text embeddings, which are used as conditional prompts to provide semantic knowledge. However, this approach lacks flexibility and cannot dynamically decode the key fine-grained semantic knowledge encoded in the latent space of MLLMs. Fine-tuning the entire model to adaptively generate Semantic Conditional Prompts (SCP) rich in fine-grained high-level semantic knowledge seems feasible, but it requires substantial resources and effort. LLamP~\cite{zheng2023large} employs LLMs to dynamically generate category features, reducing the need for extensive fine-tuning of the entire model. Nevertheless, class-specific embeddings often fail to provide useful information for individual test instances and are sensitive to class shifts. To overcome these limitations, we explore using MLLMs as conditional prompt learners to adaptively generate SCP for each instance. Specifically, decoder-only LLMs $\mathcal{D}_{N}$ are commonly employed in MLLMs, where tokens are generated autoregressively during inference, with each token being produced based on all previously generated tokens. The attention value for the current token is computed from the Key and Value derived from the previous token's attention computation. Therefore, as illustrated in Fig.~\ref{OverView}, we can cache Key $K_x$ and Value $V_x$ from the final decoder layer $\mathcal{D}_{N_{\text{last}}}$ of MLLMs, forming KV-cache~\cite{wolf2020transformers,pope2023efficiently} for each instance $I_x$:
\begin{equation}
     \mathcal{D}_{N}(Z_{x}) \xrightarrow[]{\text{offline}} (K_{x}, V_{x})
\label{eq:PG-offline}
\end{equation}
where $Z_x = \textrm{Q-former}(\textrm{ViT}(I_x))$ represents the visual representation obtained after processing $I_x$ through the ViT and Q-former structures pre-trained within the MLLMs. As shown in Fig.~\ref{fig:SCP}, during the few-shot learning phase, we construct a learnable query hidden states $P_Q$, which interacts with the offline KV-cache $(K_{x}, V_{x})$ in the final decoder layer $\mathcal{D}_{N_{\text{last}}}$ of the MLLMs. This interaction ultimately produces $P_D \in \mathbb{R}^{n \times d_{\text{mllms}}}$ adaptively through the following operations:
\begin{equation}
\begin{aligned}
    P_{\mathcal{D}} 
    &= \mathcal{D}_{N_{\text{last}}} \left[ K_{x}, V_{x}, P_{\mathcal{Q}} \right] \\
    &= L_{o} ( \left\{ K_{x}, L_{k}(P_{\mathcal{Q}}) \right\} \cdot L_{q}(P_{\mathcal{Q}}) \, \cdot \left\{ V_{x}, L_{v}(P_{\mathcal{Q}}) \right\} )
\end{aligned}
\label{eq:PG}
\end{equation}

where $n$ and $d_{\text{mllms}}$ represent the length of conditional prompt and the output dimension of $\mathcal{D}_{N_{\text{last}}}$; $L_{q}$, $L_{k}$, and $L_{v}$ are the linear layers in the multi-head self-attention mechanism for $\mathcal{D}_{N_{\text{last}}}$, which project the $P_{\mathcal{Q}}$ into the hidden state spaces of query, key, and value, respectively. The $K_{x}$ and $V_{x}$ are concatenated with the projected key and value hidden states, then matrix-multiplied ($\cdot$) with the projected query hidden state $L_{q}(P_{\mathcal{Q}})$. This product is passed through linear layer $L_{o}$ to produce the output hidden state of $\mathcal{D}_{N_{\text{last}}}$. In this process, training the entire $\mathcal{D}_{N_{\text{last}}}$ might lead to knowledge degradation, so we train the query layer $L_{q}$ to enable $P_Q$ to capture useful information from the knowledge cache of MLLMs, and update $L_{o}$ to project it into the appropriate output semantic space. $P_{\mathcal{D}}$ is mapped to the embedding space of the text encoder by introducing the learnable bottleneck layers $\mathcal{F}_{i}^{\text{t}} \in \mathbb{R}^{d_{\text{mllms}} \times d_{\text{vlms-text}}}$ at various layers of VLMs, resulting in SCP:
\begin{equation}
P_{\mathcal{TD}}^{i} = \mathcal{F}_{i}^{\text{t}} (P_{\mathcal{D}}), P_{\mathcal{TD}}^{i} \in \mathbb{R}^{n \times d_{\text{vlms-text}}}
\label{eq:PG-T-Linear}
\end{equation}

\noindent\textbf{Visual Conditional Prompts (VCP) Generation.} Most existing conditional prompt learning methods overlook the synergy between the vision and text branches, leading to a deviation of VLMs from their pre-trained knowledge. To effectively align the multi-modal space of VLMs and provide differentiated Visual Conditional Prompts (VCP) for the visual contextal prompts of different instances, we propose an Attention Mutual-Guidance (AMG) module. This module comprises two components: self-attn and cross-attn module. The self-attn component captures essential local semantic information, while the cross-attn component aggregates visual cues to map relationships between multi-modal information thereby generating VCP, enhancing SCP, and ultimately facilitating cross-layer and cross-modal instance-level prompt mutual-guidance interactions guided by textual semantics and image feature:
\begin{equation}
\begin{aligned}
P_{\mathcal{C}}^{i}= P_{\text{cross}}^{i} \circ P_{\text{self}}^{i} + \text{FFN}(\text{LN}(P_{\text{cross}}^{i} \circ P_{\text{self}}^{i}), \\
 \textit{where} \quad P_{\text{cross}}^{i} =P_{\mathcal{TD}}^{i} + \Phi_{\text{cross}}(\text{LN}(P_{\mathcal{TD}}^{i}),\; \text{LN}(F_x)), \\
 P_{\text{self}}^{i} = P_{\mathcal{TD}}^{i} + \Phi_{\text{self}}(\text{LN}(P_{\mathcal{TD}}^{i})) \\
 \end{aligned}
 \label{eq:PG-V}
\end{equation}
where $\text{LN}(\cdot)$ and $\text{FFN}(\cdot)$ represent the layer normalization and two linear layers. In this process, the key and value in the cross-attn module $\Phi_{\text{cross}}$ are derived from the VLMs image features $F_x$ of the input instance $I_x$, while the query is provided by the semantically enriched $P_{\mathcal{TD}}^{i}$. The self-attn module $\Phi_{\text{self}}$ focuses on capturing important semantic information within $P_{\mathcal{TD}}^{i}$. The outputs of these two sub-modules are enhanced through element-wise multiplication ($\circ$), which strengthens the semantic components related to image features, thereby achieving information fusion, feature filtering, and nonlinear modeling, and improving the representational capacity of $P_{\mathcal{C}}^{i} \in \mathbb{R}^{m \times d_{\text{vlms-text}}}$. Finally, the dimensions are transformed through different linear layers $\mathcal{F}_{i}^{\text{v}} \in \mathbb{R}^{d_{\text{vlms-text}} \times d_{\text{vlms-img}}}$ to obtain the VCP:
\begin{equation}
P_{\mathcal{VC}}^{i} = \mathcal{F}_{i}^{\text{v}} (P_{\mathcal{C}}^{i}), P_{\mathcal{VC}}^{i} \in \mathbb{R}^{n \times d_{\text{vlms-img}}}
\label{eq:PG-V-Linear}
\end{equation}

\begin{figure*}[t]
\centering
\includegraphics[width=0.98\textwidth]{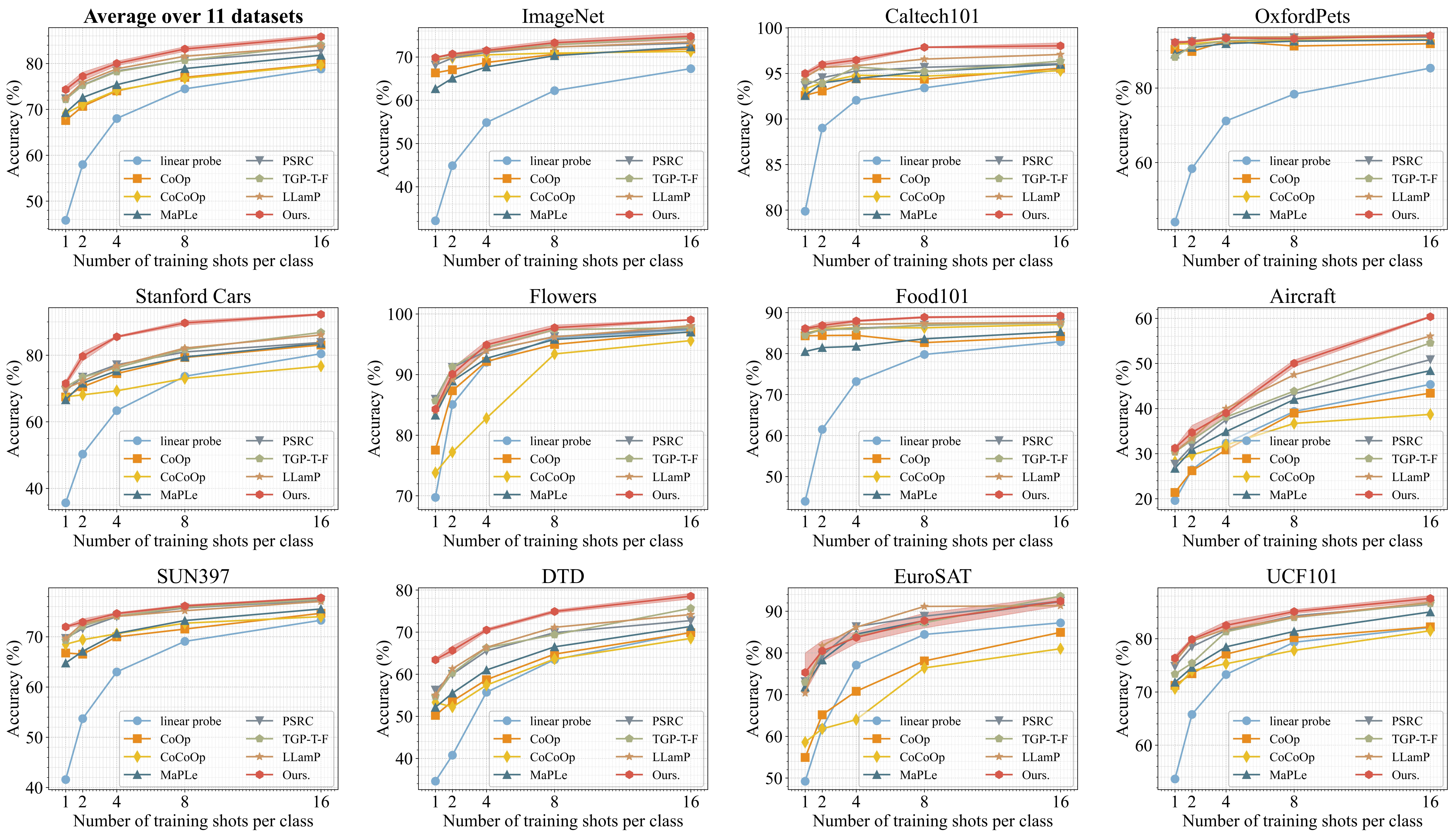}
\caption{The performance comparison of our MuGCP with the state-of-the-art methods on few-shot learning, including 1/2/4/8/16-shots on 11 datasets.}
\label{fig:few-shot}
\end{figure*}

\noindent\textbf{Multi-Prompt Fusion (MPF) Mechanism.} Directly adding conditional prompts into contextual prompts is insufficient for fully capturing the knowledge embeddings in the conditional prompts. In this paper, we propose a Multi-Prompt Fusion (MPF) mechanism that leverages multi-head self-attention in the VLMs' encoder to capture complex relationships and interactions among conditional prompts, contextual prompts, and word embeddings or patch embeddings across different attention heads, effectively fusing SCP and VCP with their corresponding text and visual contextual prompts, thereby enhancing the learning and modeling of class embeddings and instance-specific knowledge. Specifically, to merge features from different tokens provided by MLLMs and generate more informative combined feature representations, $P_{\mathcal{TD}}^{i}$ and the text contextual prompt $P_{\mathcal{T}}^{i} \in \mathbb{R}^{m \times d_{\text{vlms-text}}}$ can be extended to $\overline{P}_{\mathcal{TD}}^{i} \in \mathbb{R}^{1 \times n \times d_{\text{vlms-text}}}$ and $\overline{P}_{\mathcal{T}}^{i} \in \mathbb{R}^{m \times 1 \times d_{\text{vlms-text}}}$, respectively. The same applies to $P_{\mathcal{VC}}^{i}$ and the visual contextual prompts $P_{\mathcal{V}}^{i} \in \mathbb{R}^{m \times d_{\text{vlms-img}}}$. This allows for enhanced contextual prompts through broadcasting, summing, and averaging different tokens:
\begin{equation}
P_{\mathcal{TA}}^{i} = \frac{1}{n} \sum_{j=0}^{n} \left ( \overline{P}_{\mathcal{T}}^{i} + \overline{P}_{\mathcal{TD}}^{i} \right )
,P_{\mathcal{VA}}^{i} = \frac{1}{n} \sum_{j=0}^{n} \left ( \overline{P}_{\mathcal{V}}^{i} + \overline{P}_{\mathcal{VC}}^{i} \right )
\label{eq:PG-Fusion}
\end{equation}
We then extend the semantic conditional prompt $P_{\mathcal{TD}}^{i}$ along the class dimension and concatenate it with $P_{\mathcal{TA}}^{i} \in \mathbb{R}^{m \times d_{\text{vlms-text}}}$, word embeddings $\{{w}_l|_{l=1}^{L}\}$, start token embeddings ${w}_{SOS}$ and end token embeddings ${w}_{EOS}$. The visual conditional prompt $P_{\mathcal{VC}}^{i}$ is directly concatenated with $P_{\mathcal{VA}}^{i} \in \mathbb{R}^{m \times d_{\text{vlms-img}}}$ and image patch embeddings $E_{0} = \{v_{1}, v_{2}, ...v_{M}\}$, $M$ denotes the number of patches of image $I_x$. Consequently, Our new text embeddings $W_{i}^{\textrm{mllms}}$ and patch embeddings $E_{i}^{\textrm{mllms}}$ are represented as follows:
\begin{equation}
\begin{aligned}
& W_{i}^{\textrm{mllms}} = \{{w}_{SOS}, P_{\mathcal{TA}}^{i},{w}_{1}, \cdots, {w}_{L}, P_{\mathcal{TD}}^{i}, {w}_{EOS}\}  \\
& E_{i}^{\textrm{mllms}} = \{v_{1}, v_{2}, \cdots, v_{M}, P_{\mathcal{VC}}^{i}, P_{\mathcal{VA}}^{i}\}
\label{eq:PG-Wi-Ei}
\end{aligned}
\end{equation}
Ultimately, the image feature $F^{\ast } = \mathcal{V}_{i}(E_{i-1}^{\text{mllms}})$ and text feature $T^{\ast } = \mathcal{T}_{i}(W_{i-1}^{\text{mllms}})$ are obtained through the image encoder $\mathcal{V}_{i}$ and text encoder $\mathcal{T}_{i}$ for prediction:
\begin{equation}
\label{eq:PG-pre}
\mathcal{P} = \frac{\exp{( d(F^{\ast } \cdot T^{\ast }) / \tau)}}{\sum_{k=1}^{K}\exp{( d(F^{\ast } \cdot T^{\ast }_{j}) / \tau)}}
\end{equation}
where $\tau$ is temperature and $d(\cdot)$ is cosine similarity.

\begin{table*}[t]
\centering
\caption{Comparison with state-of-the-art methods on 16-shots base-to-new generalization. The best accuracies are bolded. HM indicates the harmonic mean.}
    \setlength{\tabcolsep}{0.01mm}
    \subfloat{
        \centering
        \begin{tabular}{lccc|ccc|ccc|ccc|ccc|ccc}
            \toprule 
            \multirow{2}[3]{*}{Method} & \multicolumn{3}{c}{Average} & \multicolumn{3}{c}{ImageNet} & \multicolumn{3}{c}{Caltech101} & \multicolumn{3}{c}{OxfordPets} & \multicolumn{3}{c}{Stanford Cars} & \multicolumn{3}{c}{Flowers} \\
            \cmidrule(lr){2-4}
            \cmidrule(lr){5-7}
            \cmidrule(lr){8-10}
            \cmidrule(lr){11-13}
            \cmidrule(lr){14-16}
            \cmidrule(lr){17-19}
            & Base & New & \multicolumn{1}{c}{HM} 
            & Base & New & \multicolumn{1}{c}{HM} 
            & Base & New & \multicolumn{1}{c}{HM}
            & Base & New & \multicolumn{1}{c}{HM}
            & Base & New & \multicolumn{1}{c}{HM}
            & Base & New & \multicolumn{1}{c}{HM}\\
            \midrule
            CLIP~\cite{radford2021learningCLIP}
            & 69.34 & 74.22 & 71.70 
            & 72.43 & 68.14 & 70.22 
            & 96.84 & {94.00} & 95.40 
            & 91.17 & 97.26 & 94.12
            & 63.37 & 74.89 & 68.65 
            & 72.08 & 77.80 & 74.83 \\
            \midrule
            CoOp~\cite{zhou2022learningCoOp}
            & 82.69 & 63.22 & 71.66 
            & {76.47} & 67.88 & 71.92
            & {98.00} & 89.81 & 93.73
            & 93.67 & 95.29 & 94.47
            & {78.12} & 60.40 & 68.13
            & {97.60} & 59.67 & 74.06
            \\
            CoCoOp~\cite{zhou2022conditionalCoCoOp}
            & 80.47 & 71.69 & 75.83 
            & 75.98 & {70.43} & {73.10}
            & 97.96 & 93.81 & {95.84}
            & {95.20} & {97.69} & {96.43}
            & 70.49 & 73.59 & {72.01}
            & 94.87 & 71.75 & {81.71}
            \\
            MaPLe~\cite{khattak2023maple}
            & 82.28 &  75.14 &  78.55
            & {76.66} &  {70.54} & {73.47}
            & 97.74 & {94.36} &  {96.02}
            &  {95.43} & {97.76} &  {96.58}
            & 72.94 & 74.00 &  {73.47}
            & 95.92 & 72.46 &  82.56
            \\
            KgCoOp~\cite{KgCoOp}
            & 80.73 & 73.60 & 77.00
            &75.83 &69.96& 72.78
            & 97.72  & 94.39  & 96.03
            & 94.65 & 97.76 & 96.18
            & 71.76  & {75.04}  & 73.36
            & 95.00  & 74.73  & 83.65
            \\
            PSRC~\cite{khattak2023self}
            & {84.26} & 76.10 & 79.97
            & 77.60 & 70.73 & 74.01
            & 98.10 & 94.03 & 96.02
            & 95.33 & 97.30 & 96.30
            & {78.27} & 74.97 & {76.58}
            & {98.07} & 76.50 & 85.95
            \\
            CoPrompt~\cite{roy2023consistency}
            & 82.91 & 75.11 & 78.58
            & 76.50 & 70.93 & 73.61
            & 98.97 & 95.50 & 97.20
            & 94.70 & 96.37 & 95.53
            & {73.10} & {70.67} & {71.86}
            & {96.97} & {75.23} & {84.73}
            \\
            TCP~\cite{yao2024tcp}
            & {83.92} & {75.13} & {79.28} 
            & 77.00 & 69.60 & 73.11
            & 98.57 & 94.70 & 96.59 
            & 94.53 & 97.10 & 95.79
            & {79.90} & {74.00} & {76.84}
            & {97.93} & {74.79} & {84.81}
            \\
            HIPL~\cite{yin2024hierarchy}
            & 83.15 & 75.14 & 78.94 
            & 76.60 & 70.40 & 73.37
            & 98.10 & 94.37 & 96.20 
            & 96.03 & 96.87 & 96.45
            & 76.73 & 70.70 & 73.59
            & 97.07 & 77.33 & 86.08
            \\
            MCPT~\cite{ren2024modality}
            & 83.92 & 75.53 & 79.61 
            & 76.98 & \bf 72.05 & 74.43
            & 98.67 & 94.63 & 96.76 
            & 95.76 & 98.23 & 96.99
            & 77.19 & 74.20 & 75.73
            & 97.20 & 72.87 & 83.37
            \\
            LLamP~\cite{zheng2023large}
            & 84.95 & 76.76 & 80.65
            & 78.00 & 70.88 & 74.27
            & 98.92 & 95.70 & 97.29
            & 95.82 & 97.54 & 96.67
            & 81.22 & 73.96 & 77.42
            & 97.91 & 76.99 & 86.20
            \\
            \midrule
            \rowcolor{tabhighlight}
            & \bf 87.08 & \bf 77.53 & \bf 82.03
            & \bf 79.07 & 70.43 & \bf 74.49
            & \bf 99.00 & \bf 95.70 & \bf 97.32
            & \bf 96.00 & \bf 97.76 & \bf 96.84
            & \bf 89.00 & \bf 77.66 & \bf 82.94
            & \bf 99.07 & \bf 77.60 & \bf 87.03
            \\
            \rowcolor{tabhighlight}
        \multirow{-2}{*}{Ours (w/ Std)}
        & \scriptsize ($\pm0.16 $) & \scriptsize ($\pm0.42 $) & \scriptsize ($\pm0.31 $)
        & \scriptsize ($\pm1.32 $) & \scriptsize ($\pm0.26 $) & \scriptsize ($\pm0.56 $)
        & \scriptsize ($\pm0.16 $) & \scriptsize ($\pm0.36 $) & \scriptsize ($\pm0.19 $)
        & \scriptsize ($\pm0.36 $) & \scriptsize ($\pm0.43 $) & \scriptsize ($\pm0.15 $)
        & \scriptsize ($\pm0.43 $) & \scriptsize ($\pm0.69 $) & \scriptsize ($\pm 0.58$)
        & \scriptsize ($\pm0.21 $) & \scriptsize ($\pm0.46 $) & \scriptsize ($\pm0.25 $)
        \\
            % \bottomrule
        \end{tabular}
    % \end{subtable}
    }
    \vspace*{-0.3cm}
    \\
    % \begin{subtable}[t]{\textwidth}
    \subfloat{
        \centering
        % \vspace{5pt}
        \begin{tabular}{lccc|ccc|ccc|ccc|ccc|ccc}
            \toprule 
            \multirow{2}[3]{*}{Method} & \multicolumn{3}{c}{Food101} & \multicolumn{3}{c}{Aircraft}& \multicolumn{3}{c}{SUN397} & \multicolumn{3}{c}{DTD} & \multicolumn{3}{c}{EuroSAT} & \multicolumn{3}{c}{UCF101} \\
            \cmidrule(lr){2-4}
            \cmidrule(lr){5-7}
            \cmidrule(lr){8-10}
            \cmidrule(lr){11-13}
            \cmidrule(lr){14-16}
            \cmidrule(lr){17-19}
            & Base & New & \multicolumn{1}{c}{HM} 
            & Base & New & \multicolumn{1}{c}{HM} 
            & Base & New & \multicolumn{1}{c}{HM} 
            & Base & New & \multicolumn{1}{c}{HM}
            & Base & New & \multicolumn{1}{c}{HM}
            & Base & New & \multicolumn{1}{c}{HM}\\
            \midrule
            CLIP~\cite{radford2021learningCLIP}
            & 90.10 & 91.22 & 90.66 
            & 27.19 & 36.29 & 31.09
            & 69.36 & 75.35 & 72.23 
            & 53.24 & 59.90 & 56.37 
            & 56.48 & 64.05 & 60.03 
            & 70.53 & 77.50 & 73.85\\
            \midrule
            CoOp~\cite{zhou2022learningCoOp}
            & 88.33 & 82.26 & 85.19
            & {40.44} & 22.30 & 28.75
            & {80.60} & 65.89 & 72.51
            & {79.44} & 41.18 & 54.24
            & {92.19} & 54.74 & 68.69
            & {84.69} & 56.05 & 67.46            
             \\ 
            CoCoOp~\cite{zhou2022conditionalCoCoOp}
            & {90.70} & {91.29} & {90.99}
            & 33.41 & 23.71 & 27.74
            & 79.74 & {76.86} & {78.27}
            & 77.01 & 56.00 & {64.85}
            & 87.49 & 60.04 & {71.21}
            & 82.33 & 73.45 & {77.64}            
             \\
            MaPLe~\cite{khattak2023maple}
            &  {90.71} &  92.05 &  91.38
            & 37.44 & 35.61 & {36.50}
            & {80.82} &  {78.70} &  {79.75}
            & {80.36} & 59.18 &  {68.16} 
            & {94.07} &  {73.23} & {82.35}
            & 83.00 &  {78.66} & {80.77}            
             \\
            KgCoOp~\cite{KgCoOp}
            & 90.50 &  91.70 & 91.09
            & 36.21 & 33.55 & 34.83
            & 80.29 & 76.53 & 78.36
            & 77.55 & 54.99 & 64.35
            & 85.64 & 64.34 & 73.48
            & 82.89 & 76.67  & 79.65            
             \\
            PSRC~\cite{khattak2023self}
            & 90.67 & 91.53 & 91.10
            & {42.73} & 37.87 & {40.15}
            & {82.67} & 78.47 & 80.52
            & {83.37} & 62.97 & 71.75
            & 92.90 & 73.90 & 82.32
            & {87.10} & 78.80 & 82.74            
             \\
            CoPrompt~\cite{roy2023consistency}
            & 90.27 & 91.83 & 91.04
            & {36.50} & 33.50 & 34.94
            & {82.33} & {79.40} & {80.84}
            & {82.77} & {61.27} & {70.42}
            & {93.47} & {73.20} & {82.10}
            & {86.30} & {78.27} & {82.09}            
             \\
            TCP~\cite{yao2024tcp}
            & 90.67 & 91.40 & 91.03
            & {42.05} & 34.33 & 37.79
            & {82.67} & {78.17} & {80.36}
            & {82.67} & {56.87} & {67.38}
            & {90.03} & {74.50} & {81.53}
            & {87.13} & \bf{80.93} & \bf{83.92}            
            \\
            HIPL~\cite{yin2024hierarchy}
            & 90.30 & 91.33 & 90.81
            & 37.50 & 34.20 & 35.77
            & 80.70 & 77.93 & 79.29
            & 81.57 & 52.73 & 64.05
            & \bf 96.33 & \bf 83.70 & \bf 89.55
            & 83.67 & 76.97 & 80.16            
            \\
            MCPT~\cite{ren2024modality}
            & 90.75 & \bf 92.10 & 91.46
            & 39.10 & 36.50 & 37.83
            & 81.69 & 78.80 & 80.27
            & 84.10 & 59.56 & 71.86
            & 95.43 & 73.77 & 84.09
            & 86.83 & 78.73 & 82.81            
            \\
            LLamP~\cite{zheng2023large}
            & 90.51 & 91.81 & 91.16
            & 48.34 & 36.93 & 41.87
            & 83.10 & 79.36 & 81.19
            & 82.53 & 63.12 & 71.53
            & 90.65 & 79.16 & 84.52
            & \bf 87.42 & 78.92 & 82.95            
             \\
            \midrule
            \rowcolor{tabhighlight}
            & \bf 91.77 & 91.37 & \bf 91.57
            & \bf 52.10 & \bf 39.11 & \bf 44.68
            & \bf 84.23 & \bf 79.41 & \bf 81.75
            & \bf 85.10 & \bf 64.77 & \bf 73.56
            &  95.56 & 78.36 & 86.11
            & 87.03 & 80.70 & 83.75            
             \\
            \rowcolor{tabhighlight}
        \multirow{-2}{*}{Ours (w/ Std)}
        & \scriptsize ($\pm0.05 $) & \scriptsize ($\pm0.12 $) & \scriptsize ($\pm0.07 $)
        & \scriptsize ($\pm0.88 $) & \scriptsize ($\pm1.91 $) & \scriptsize ($\pm1.11$)
        & \scriptsize ($\pm0.16 $) & \scriptsize ($\pm0.42 $) & \scriptsize ($\pm0.31 $)
        & \scriptsize ($\pm0.85 $) & \scriptsize ($\pm1.16 $) & \scriptsize ($\pm0.94$)
        & \scriptsize ($\pm0.68 $) & \scriptsize ($\pm1.93 $) & \scriptsize ($\pm1.44 $)
        & \scriptsize ($\pm0.37 $) & \scriptsize ($\pm0.94 $) & \scriptsize ($\pm0.65 $))
        \\
            \bottomrule
        \end{tabular}
    % \end{subtable}
    }
    \\
    \label{table:16-shots base-to-new}
\end{table*}

\subsection{Text Augmentations.}
Enforcing consistency supervision between the trained model and the pre-trained model can encourage the general knowledge within pre-trained models for prompt learning, thereby preventing overfitting and spatial drift~\cite{khattak2023self,roy2023consistency}. In text supervision, we aim to fully leverage MLLMs $\mathcal{H}^{\textrm{mllms}}$ to extract general and broad high-level visual information $T^{\textrm{mllms}}_{K} = \mathcal{H}^{\textrm{mllms}}(I^{\textrm{train}}_{K})$ for each class $K$ from the training instances $I^{\textrm{train}}$. Furthermore, the diversity of text supervision is extended in two ways: (1) visual custom text $D^{\textrm{mllms}}_{K}$: By relying on our customized hand-crafted template ``a photo of a $\left \{\textrm{[class name]}_K\right \}$, with $\left \{T^{\textrm{mllms}}_{K}\right \}$.'' to embed class names and high-level visual information, thereby enriching content descriptions; (2) visual expansion text $D^{\textrm{llm}}_{K}$: By using $T^{\textrm{mllms}}_{K}$ to construct prompt templates $\rho^{\textrm{llm}}$ for the GPT-3.5~\cite{gpt-4}, guiding it to generate more detailed and realistic text descriptions $D^{\textrm{llm}}_{K} = \textrm{GPT-3.5} \left \{ \rho^{\textrm{llm}}(T^{\textrm{mllms}}_{K}) \right \}$. Specifically, these texts use visual information to distinguish themselves from other categories within the same class. Our query LLMs prompt templates $\rho^{\textrm{llm}}$ can be divided into the following three parts: i) we embed ``[class-names]'' and ``[domain-names]'' to prompt the LLMs to generate discriminative descriptions related to the dataset categories; ii) we embed the visual information $T^{\textrm{mllms}}_{K}$ to guide the LLMs; iii) we ultimately obtain a fixed format output $D^{\textrm{llm}}_{K}$. the $\rho^{\textrm{llm}}$ as follows:
\begin{center}
\colorbox{cyan!15}{%
  \parbox{0.46\textwidth}{%
  \centering
    What fine-grained characteristics can be used to differentiate the \textbf{[class name]$\mathbf{_K}$} and \textbf{[domain-names]}? Combine the VQA-description: [$\mathbf{T^{\textrm{mllms}}_{K}}$].
    Texts should be of the form: ``<A/An/The> <category> is <VQA-description> and <fine-grained characteristics>''.
  }%
}
\end{center}

\subsection{Loss Function}
Consistent with CoPrompt~\cite{roy2023consistency}, we employed the cross-entropy loss $\mathcal{L}_{ce}$ and the consistency loss $\mathcal{L}_{cc} = \mathcal{L}_{text} + \mathcal{L}_{img}$ to supervise training:
\begin{equation}
\begin{aligned}
 & \mathcal{L} = \mathcal{L}_{ce} + \lambda \mathcal{L}_{cc} \\
 \textit{where} \quad &\mathcal{L}_{ce} = - log \frac{\text{exp}(d(\Phi_{\text{va}}(F^{\ast }), \Phi_{\text{ta}}(T^{\ast }))/\tau)}{\sum_{k=1}^{K}\text{exp}(d(\Phi_{\text{va}}(F^{\ast }),  \Phi_{\text{ta}}(T^{\ast })_{k})/\tau)}, \\
 &\mathcal{L}_{cc} = 2 - \frac{T' \cdot \Phi_{\text{ta}}(T^{\ast })}{\left \| T' \right \| \left \| \Phi_{\text{ta}}(T^{\ast }) \right \|} - \frac{F' \cdot \Phi_{\text{va}}(F^{\ast })}{\left \| F' \right \| \left \| \Phi_{\text{va}}(F^{\ast }) \right \|} \\
 \end{aligned}
 \label{eq:ce}
\end{equation}
where $\lambda$ is a loss balance factor; $T'$ denotes $T' = \mathcal{T}_{i}(W^{\text{aug}}_{i-1})$; $W^{\text{aug}}_{0}$ signifies the text embeddings of the augmented text $(D^{\textrm{mllms}}_{K} + D^{\textrm{llm}}_{K})$; $F'$ represents the randomly augmented image features; $\Phi_{\text{va}}$ and $\Phi_{\text{ta}}$ are the learnable image and text adapters.

\section{Experiment}
\subsection{Experimental Settings}
\noindent\textbf{Datasets.} To ensure a fair and impartial evaluation of our MuGCP. We evaluate MuGCP on 14 different datasets. These include generic object datasets such as ImageNet~\cite{deng2009imagenet} and Caltech101~\cite{fei2004learningCaltech101}; fine-grained classification datasets like Oxford Pets~\cite{parkhi2012catsOxfordPets}, Stanford Cars~\cite{krause20133dStandfordCars}, Flowers~\cite{nilsback2008automatedFlowers102}, Food101~\cite{bossard2014foodFood101}, and Aircraft~\cite{maji2013fineFGVCAircraft}; the scene recognition dataset SUN397~\cite{xiao2010sunSUN397}; the action recognition dataset UCF101~\cite{soomro2012ucf101}; the texture classification dataset DTD~\cite{cimpoi2014describingDTD}; the satellite imagery dataset EuroSAT~\cite{helber2019eurosatEuroSAT}; and three ImageNet variants, namely ImageNetV2~\cite{recht2019imagenet}, ImageNet-Sketch~\cite{wang2019learning}, and ImageNet-A~\cite{hendrycks2021natural}.

\noindent\textbf{Evaluation protocols.} Following the work of ~\cite{zhou2022conditionalCoCoOp,zhou2022learningCoOp,zheng2023large,khattak2023self,yao2024tcp}, we evaluate the effectiveness from the four scenarios: (1) few-shot learning with 1/2/4/8/16-shots labeled images; (2) 1/2/4/8/16-shots base-to-new generalization; (3) cross-dataset generalization from the ImageNet to 11 other different datasets; (4) domain generalization from the ImageNet to three other ImageNet variant datasets.

\noindent\textbf{Baselines.} (1) Domain-shared prompt learning: CoOp~\cite{zhou2022learningCoOp}, MaPLe~\cite{khattak2023maple}, KgCoOp~\cite{KgCoOp}, PSRC~\cite{khattak2023self}, CoPrompt~\cite{roy2023consistency}, TCP~\cite{yao2024tcp}, TGP-T-F~\cite{tan2024compound}, HIPL~\cite{yin2024hierarchy}, MCPT~\cite{ren2024modality} and LLamP~\cite{zheng2023large}. (2) Conditional prompt learning: CoCoOp~\cite{zhou2022conditionalCoCoOp}.

\noindent\textbf{Implementation Details.} We employed CLIP (ViT-B/16)~\cite{radford2021learningCLIP} as the pre-trained vision language model backbone to evaluate our MuGCP. MiniGPT-4 (llama2-chat-7b)~\cite{zhu2023minigpt} and BLIP-2 (opt-2.7b)~\cite{li2023blip} served as conditional prompt learners, utilizing a conditional prompt length of $n=16$. All experiments were conducted on a single RTX 3090 GPU. For the experimental setup across the four evaluation scenarios, both text and visual contextual prompts were learned across transformer layers $l=12$ with a contextual prompt length of $m=4$. The text prompts in the first layer was initialized with the word embeddings ``X X X X $\left \{  \right \}$ '', while all other prompts were randomly initialized from a normal distribution. To ensure a fair comparison, the final performance was averaged over three random seeds (1/2/3) and the Standard deviation (Std) was reported. We used AdamW optimizer~\cite{loshchilov2017decoupled} with a learning rate of 2e-4 and a batch size of 1. For few-shot learning, our MuGCP is trained for 20 epochs, while the base-to-new setting is trained for 8 epochs.

\begin{table*}[t]
\caption{For 1/2/4/8/16-shots base to new generalization on 11 datasets, the average results across the 11 datasets are shown. `*' indicates that we use the same three seeds to reimplement performance for fair comparison. }
    % \small
    % \renewcommand\tabcolsep{5pt}
    % \scriptsize
    \begin{center}
    \setlength{\tabcolsep}{0.58mm}
    \begin{tabular}
    {lccc|ccc|ccc|ccc|ccc}
    % {lccc|ccc}
    \toprule
    & \multicolumn{3}{c}{1-shot} 
    & \multicolumn{3}{c}{2-shots} 
    & \multicolumn{3}{c}{4-shots} 
    & \multicolumn{3}{c}{8-shots}  
    & \multicolumn{3}{c}{16-shots} 
    \\
    Method
    & Base & New & HM 
    & Base & New & HM 
    & Base & New & HM 
    & Base & New & HM 
    & Base & New & HM
    \\
    \cmidrule(r){1-1}
    \cmidrule(r){2-4}
    \cmidrule(r){5-7}
    \cmidrule(r){8-10}
    \cmidrule(r){11-13}
    \cmidrule(r){14-16}
     CLIP~\cite{radford2021learningCLIP}
        & 69.34 & 74.22 &  71.70
        & 69.34 & 74.22 &  71.70
        & 69.34 & 74.22 &  71.70
        & 69.34 & 74.22 &  71.70
        & 69.34 & 74.22 &  71.70
        \\   
    \cmidrule(r){1-1}
    \cmidrule(r){2-4}
    \cmidrule(r){5-7}
    \cmidrule(r){8-10}
    \cmidrule(r){11-13}
    \cmidrule(r){14-16}
    CoOP*~\cite{zhou2022learningCoOp}
        & 70.71  & 64.98  & 67.73
        & 74.52  & 63.09  & 68.33
        & 77.52  & 64.07  & 70.15
        & 80.32  & 64.18  & 71.35
        & 82.69  & 63.22 & 71.66
        \\
    CoCoOp*~\cite{zhou2022conditionalCoCoOp}
        &  72.07 & 71.80  & 71.93
        &  73.93 & 71.94  & 72.92
        &  76.64 & 72.39  & 74.46
        &  78.56 & 72.51  & 75.41
        &  80.47 & 71.69  & 75.83
        \\      
    MaPLe*~\cite{khattak2023maple}
        &  71.76 &  71.78 &  71.77
        &  74.84 &  73.84 &  74.33
        &  77.20 &  73.16 &  75.13
        &  79.98 &  73.45 &  76.57
        &  82.28 &  75.14 &  78.55
        \\
    KgCoOp*~\cite{KgCoOp}
        &  75.48 & 73.16  & 74.30
        &  76.89 & 73.67  & 75.25
        &  77.99 & 73.74  & 75.80
        &  79.42 & 73.54  & 76.37
        &  80.73 & 73.60 & 77.00
        \\       
    PSRC*~\cite{khattak2023self}
        &  73.93 &  71.38 & 72.63
        &  77.95 &  73.07 & 75.43
        &  80.43 &  75.47 & 77.87
        &  82.12 &  74.66 & 78.21
        &  84.26 & 76.10 & 79.97
        \\      
    CoPrompt*~\cite{roy2023consistency}
        &  70.63 &  70.14 & 70.38
        &  74.22 &  71.87 & 73.02
        &  78.39 &  75.17 & 76.74
        &  80.52 &  74.41 & 77.34
        &  82.91 & 75.11 & 78.58
        \\
    TCP*~\cite{yao2024tcp}
        &  75.51 &  74.65 & 75.08
        &  77.98 &  75.01 & 76.47
        &  80.25 &  74.71 & 77.39
        &  81.81 &  74.90 & 78.21
        &  83.92 & 75.13 & 79.28
        \\  
    LLamP*~\cite{zheng2023large}
        &  75.17 &  75.21 & 75.18
        &  78.46 &  75.43 & 76.92
        &  80.49 &  75.38 & 77.85
        &  82.52 &  76.46 & 79.37
        &  84.95 & 76.76 & 80.65
        \\   
    \cmidrule(r){1-1}
    \cmidrule(r){2-4}
    \cmidrule(r){5-7}
    \cmidrule(r){8-10}
    \cmidrule(r){11-13}
    \cmidrule(r){14-16}
     \rowcolor{tabhighlight}
        &  \bf 76.82 & \bf 75.69  & \bf 76.25
        &  \bf 79.27 & \bf 76.24  & \bf 77.72
        &  \bf 81.36 & \bf 75.93  & \bf 78.55
        &  \bf 84.05 & \bf 76.79  & \bf 80.26 
        &  \bf 87.08 & \bf 77.53  & \bf 82.03
        \\
        \rowcolor{tabhighlight}
        \multirow{-2}{*}{Ours (w/ Std)}
        &  ($\pm0.23 $) &  ($\pm0.33 $) & ($\pm0.29 $)
        &  ($\pm0.52 $) &  ($\pm0.43 $) & ($\pm0.38 $)
        &  ($\pm0.47 $) &  ($\pm0.35 $) & ($\pm0.41 $)
        &  ($\pm0.43 $) &  ($\pm0.81$) &  ($\pm0.62$)
        &  ($\pm0.37 $) &  ($\pm0.94 $) & ($\pm0.65 $) 
        \\
    \bottomrule
    \end{tabular}
    % }
    \end{center}
    \label{tab:1/2/4/8/16-shots base-to-new}

\end{table*}

\subsection{Comparison with State-of-the-Art Methods}
\noindent\textbf{Base-to-new generalization.} In the 1/2/4/8/16-shots base-to-new generalization settings, MuGCP consistently achieves state-of-the-art performance in the Harmonic Mean (HM), as detailed in Table~\ref{tab:1/2/4/8/16-shots base-to-new}. In the 16-shots setting, MuGCP shows an average improvement of 2.13\% in base classes accuracy, 0.65\% in new classes accuracy, and 1.31\% in HM compared to LLamP. These gains are also evident in the 1/2/4/8-shots settings, with MuGCP's 8-shots performance matching LLamP's 16-shots results, highlighting its effectiveness in learning from limited data. MuGCP consistently outperforms other prompt learning methods in HM, reflecting a superior balance between base and new classes generalization. Notably, as shown in Table~\ref{table:16-shots base-to-new}, in the 16-shots setting on the Stanford Cars dataset, MuGCP surpasses LLamP by 3.70\% in new classes accuracy and 7.78\% in base classes accuracy, with similar improvements observed on the Aircraft dataset. These results underscore MuGCP's ability to leverage fine-grained high-level semantic knowledge provided by SCP and VCP effectively, capturing the most discriminative features for fine-grained object classification.

\begin{table}[h]
    \centering
    \caption{The few-shot Learning average results of various methods on 11 datasets.}
    \label{tab:few-shot}
    \setlength{\tabcolsep}{1mm}
    \begin{tabular}{c|ccccc}
        \toprule
        Method & 1-shot & 2-shots & 4-shots & 8-shots & 16-shots\\
        \midrule
        Linear probe & 45.83 & 57.98 & 68.01 & 74.47 & 78.79  \\
        CoOp~\cite{zhou2022learningCoOp} & 67.59 & 70.65 & 74.02 & 76.98 & 79.89  \\
        CoCoOp~\cite{zhou2022conditionalCoCoOp} & 69.10 & 71.02 & 74.14 & 76.78 & 79.69  \\
        MaPLe~\cite{khattak2023maple} & 69.27 & 72.58 & 75.37 & 78.89 & 81.79  \\
        PSRC~\cite{khattak2023self} & 72.32 & 75.29 & 78.35 & 80.69 & 82.87  \\
        TGP-T-F~\cite{tan2024compound} & 72.15 & 75.22 & 78.20 & 80.69 & 84.06 \\
        HIPL~\cite{yin2024hierarchy} & 71.59 & 74.43 & 77.41 & 80.34 & 82.61  \\
        LLamP~\cite{zheng2023large} & 72.42 & 75.89 & 78.83 & 81.57 & 83.81  \\
        \rowcolor{tabhighlight} & \bf 74.31 & \bf 77.21 & \bf 80.01 & \bf 83.15 & \bf 85.82 \\
        \rowcolor{tabhighlight} \multirow{-2}{*}{Ours (w/ Std)}& { ($\pm0.76 $)}  & { ($\pm0.88 $)}  & { ($\pm0.50 $)} & { ($\pm0.50 $)} & { ($\pm0.42 $)} \\    
    \bottomrule
    \end{tabular}
\end{table}
\noindent\textbf{Few-shot learning.} To evaluate the superiority and reliability of MuGCP in learning task-specific high-level semantic knowledge, we compare it with existing prompt learning methods. As shown in Fig.~\ref{fig:few-shot} and Table~\ref{tab:few-shot}, MuGCP consistently outperforms previous prompt learning methods in average performance across 11 datasets for 1/2/4/8/16-shots settings. Notably, in the 16-shots setting, MuGCP achieves an average performance of 85.82\%, surpassing LLamP by 2.01\% and TGP-T-F by 1.76\%. Additionally, MuGCP demonstrates significant advantages on the challenging fine-grained datasets (e.g., Aircraft and Staford Cars datasets), across all shots settings. The Std analysis further confirms the robustness of MuGCP, showing superior results for each seed compared to previous methods, except for EuroSAT. These results indicate that MuGCP effectively leverages high-level semantic knowledge provided by our conditional prompts and integrates contextual prompts, thereby enhancing the acquisition of task-specific knowledge.

\begin{table*}[t]
  \centering
  \caption{The performance on cross-dataset generalization. The results superior to TCP~\cite{yao2024tcp} are in bold.}
\setlength{\tabcolsep}{1.5mm}
  \begin{tabular}{lccccccccccc}
    \toprule
    & \multicolumn{1}{c}{\bf Source} & \multicolumn{10}{c}{\bf Target} \\ \cmidrule(lr){2-2} \cmidrule(lr){3-12}
     Method
     & \footnotesize\rotatebox{0}{ImageNet} 
     & \footnotesize\rotatebox{0}{Caltech101} 
     & \footnotesize\rotatebox{0}{OxfordPets} 
     & \footnotesize\rotatebox{0}{Stanford Cars} 
     & \footnotesize\rotatebox{0}{Flowers} 
     & \footnotesize\rotatebox{0}{Food101} 
     & \footnotesize\rotatebox{0}{Aircraft} 
     & \footnotesize\rotatebox{0}{SUN397} 
     & \footnotesize\rotatebox{0}{DTD} 
     & \footnotesize\rotatebox{0}{EuroSAT} 
     & \footnotesize\rotatebox{0}{UCF101} 
     \\
    \midrule
    CoOP~\cite{zhou2022learningCoOp} & 71.51 & 93.70 & 89.14 & {65.51} & 68.71 & 85.30 & 18.47 & 64.15 & 41.92 & 46.39 & 66.55 \\
    CoCoOp~\cite{zhou2022conditionalCoCoOp} & 71.02 & {94.43} & 90.14 & 65.32 & 71.88 & {86.06} & 22.94 & 67.36 & 45.73 & 45.37 & 68.21 \\
    MaPLe~\cite{khattak2023maple} & 70.72 & 93.53 & {90.49} & {65.57} & {72.23} & {86.20} & {24.74} & 67.01 & {46.49} & {48.06} & {68.69} \\
    PSRC~\cite{khattak2023self} & 71.27 & 93.60 & 90.25 & {65.70} & 70.25 & 86.15 & 23.90 & 67.10 & 46.87 & 45.50 & 68.75 \\
    CoPrompt~\cite{roy2023consistency} &  70.80 & {94.50}& {90.73} & {65.67} & {72.30} & {86.43} & {24.00} & {67.57} & {47.07} & {51.90} & {69.73} \\
    HIPL~\cite{yin2024hierarchy} &  71.55 & 94.73& 91.10 & 63.30 & 72.03 & {86.17} & 22.17 & {63.90} & {43.47} & {54.73} & {69.33} \\
    TCP~\cite{yao2024tcp} &  71.40 & {93.97}& {91.25} & {64.69} & {71.21} & {86.69} & {23.45} & {67.15} & {44.35} & {51.45} & {68.73} \\
    % \rowcolor{tabhighlight}
    \midrule
     \rowcolor{tabhighlight}
    & \bf 74.80 & \bf 95.00 & 90.20 & \bf 65.43 & 70.10 & 85.30 & \bf 24.00 & \bf  67.17 & \bf  45.70 & 46.60 & \bf 69.33 \\
    \rowcolor{tabhighlight}
    \multirow{-2}{*}{Ours (w/ Std)}
    &  ($\pm0.75 $) &  ($\pm0.41 $) &  ($\pm1.12 $) &  ($\pm0.24 $) &  ($\pm0.58 $) &  ($\pm0.54 $) &  ($\pm0.33 $) &  ($\pm0.37 $) &  ($\pm0.34 $) & ($\pm3.23 $) & ($\pm0.91 $) \\
  \bottomrule
  \end{tabular}
  % }
  \label{tab:cross-dataset}
\end{table*}

\noindent\textbf{Cross-dataset generalization.} As illustrated in Table~\ref{tab:cross-dataset}, our model is trained on 16-shots ImageNet and evaluated on 10 unrelated datasets, demonstrating the cross-dataset generalization capability of MuGCP. MuGCP outperforms CoCoOp on 7 datasets and surpasses the state-of-the-art TCP method on 6 datasets. Notably, MuGCP achieves superior generalization on Caltech101 and UCF101, showcasing its ability to effectively leverage fine-grained high-level semantic knowledge from multi-modal conditional prompts to enhance contextual prompts in downstream task distributions.

\begin{table}[ht]
\centering
\setlength{\tabcolsep}{2.5mm}
\caption{The performance on domain generalization.}
\begin{tabular}{l cccc}
    \toprule
    & \textbf{Source} & \multicolumn{3}{c}{\textbf{Target}} \\ \cmidrule(lr){2-2} \cmidrule(lr){3-5}
    Method & ImageNet & -V2 & -Sketch & -A \\
    \midrule
    CLIP~\cite{radford2021learningCLIP} &  66.73 & 60.83 & {46.15} & 47.77 \\
    CoOp~\cite{zhou2022learningCoOp} &  {71.51} & 64.20 & 47.99  & 49.71  \\
    CoCoOp~\cite{zhou2022conditionalCoCoOp} & 71.02 & {64.07} & 48.75 & 50.63  \\
    KgCoOp~\cite{KgCoOp} & 71.20 & 64.10 & 48.97 & 50.69  \\ 
    MaPLe~\cite{khattak2023maple} & 70.72 & {64.07} & {49.15} & {50.90} \\
    PSRC~\cite{khattak2023self} & 71.27 & {64.35} & {49.55} & 50.90 \\
    CoPrompt~\cite{roy2023consistency} & {70.80} & {64.25} & {49.43} & 50.50 \\
    HIPL~\cite{yin2024hierarchy} & 72.20 & 64.80 & 48.20 & 48.50 \\
    MCPT~\cite{ren2024modality} & 72.13 & 64.73 & 49.87 & 51.03 \\
    TCP~\cite{yao2024tcp} &  71.20 & 64.60 & 49.50 & 51.20 \\
    TGP-T-F~\cite{tan2024compound} & {73.50} & {65.10} & {48.70} & / \\
    \midrule
    % \rowcolor{tabhighlight} 
    \rowcolor{tabhighlight}
    & \bf{74.80} & \bf 67.50 & \bf 53.63 & \bf 52.30 \\
    \rowcolor{tabhighlight}
    \multirow{-2}{*}{Ours (w/ Std)}
    & ($\pm0.75 $) & ($\pm1.39 $) & ($\pm3.52 $) & ($\pm 0.80$) \\
    \bottomrule
    \end{tabular}
\label{tab:domain_generalization}
\end{table}
\noindent\textbf{Domain generalization.} In this setting, we train our MuGCP on 16-shots ImageNet and test on out-of-domain (OOD) datasets, including ImageNetV2, ImageNet-Sketch, and ImageNet-A, demonstrating the model's robustness in generalizing to target domains beyond the training set. As shown in Table~\ref{tab:domain_generalization}, MuGCP achieves the best performance on ImageNet and outperforms all existing methods on the OOD datasets. These results demonstrate that MuGCP's fine-grained high-level semantic knowledge significantly enhances the robustness and generalization of contextual prompts across domain shifts.

\subsection{Ablation Study}
We conducted ablation studies to verify the effectiveness of the proposed module in our  MuGCP by evaluating 16-shots base-to-new generalization across 11 datasets.

\begin{table}[ht]
    \centering
    \caption{Ablation Study on MLLMs Knowledge. \xmark{} refers to ``w/o MLLMs, only initialize $P_{\mathcal{D}}$ with a normal distribution''. $\ddagger$ indicates replacing the image-conditional prompts in CoCoOp~\cite{zhou2022conditionalCoCoOp} with our semantic conditional prompts (SCP) while keeping all content and experimental settings consistent with CoCoOp.}
    \setlength{\tabcolsep}{3.5mm}
    \begin{tabular}{c|c|cc|c}
        \toprule
        Method & MLLMs & Base & New & HM \\
        \midrule
        CoCoOp~\cite{zhou2022conditionalCoCoOp} & / & 80.47 & 71.69 & 75.83 \\
        $\text{CoCoOp}^{\ddagger}$ & BLIP-2 & \bf 80.96 & 73.40 & 77.00 \\
        $\text{CoCoOp}^{\ddagger}$ & MiniGPT-4 & 80.64 & \bf 73.73 & \bf 77.03 \\
        \midrule
        \multirow{3}{*}{Ours}
        & \xmark & 84.01 & 76.09 & 79.85 \\
        & BLIP-2 & \bf 87.36 & 77.11 & 81.91 \\   
        & MiniGPT-4 
        & \cellcolor{tabhighlight}     87.08
        & \cellcolor{tabhighlight} \bf 77.53 
        & \cellcolor{tabhighlight} \bf 82.03
        \\
    \bottomrule
    \end{tabular}
    \label{tab:abl-vqamodel}
\end{table}

\noindent\textbf{Effect of MLLMs Knowledge.} We investigated the effectiveness of adaptively generated conditional prompts in MuGCP by MLLMs and evaluated compatibility with different MLLMs. Table~\ref{tab:abl-vqamodel} shows the performance of $\text{CoCoOp}^{\ddagger}$, which leverages MLLMs such as MiniGPT-4~\cite{zhu2023minigpt} or BLIP-2~\cite{li2023blip} to dynamically generate SCP, replacing the image-conditioned prompts used in CoCoOp~\cite{zhou2022conditionalCoCoOp}.  It can be observed that $\text{CoCoOp}^{\ddagger}$ improves the new classes accuracy by 2.04\% (with MiniGPT-4) while maintaining the base classes accuracy. This indicates that the SCP generated by our approach provides more high-level semantic knowledge and exhibits superior generalization capability. Similarly, in our MuGCP, the method of dynamically generating SCP using MLLMs significantly outperforms the approach of randomly initialized SCP in terms of both base classes accuracy and generalization to new classes. Furthermore, MuGCP demonstrates strong compatibility with various mainstream MLLMs (e.g., MiniGPT-4, BLIP-2), all of which achieve competitive results (see Table~\ref{tab:abl-vqamodel}).

\begin{table}[h]
    \centering
    \caption{Ablation study on AMG. \xmark{} refers to ``a single linear layer''. 
    }
    \setlength{\tabcolsep}{2mm}
    \begin{tabular}{c|cc|cc|c}
        \toprule
        AMG module & self-attn & cross-attn & Base & New & HM \\
        \midrule
        \multirow{4}{*}{shared AMG}
        & \xmark & \xmark & 85.87 & 75.43 & 80.31 \\
        & \cmark & & 86.65 & 75.55 & 80.72 \\
        & & \cmark & 86.39 & 76.11 & 80.93 \\
        & \cmark & \cmark & \cellcolor{tabhighlight} \bf 87.08 & \cellcolor{tabhighlight} \bf 77.53 & \cellcolor{tabhighlight} \bf 82.03 \\
        \midrule
        non-shared AMG & \cmark & \cmark & 86.40 & 76.10	& 80.89 \\
    \bottomrule
    \end{tabular}
    % }
    \label{tab:abl-module}  
\end{table}
\noindent\textbf{Effect of AMG Module.} We evaluated various modules for cross-layer and cross-modal connections in the multi-modal space of VLMs, including cross-attn, self-attn, and a linear layer. As shown in Table~\ref{tab:abl-module}, using a linear layer alone does not improve performance for both base and new classes. The self-attn module captures essential local semantic information but overfits the base classes, limiting its generalization ability. In contrast, the cross-attn module aggregates visual cues to map relationships between multi-modal information, thereby enhancing generalization to new classes. The outputs of the self-attn and cross-attn modules are combined through element-wise multiplication, enabling information fusion, feature filtering, and nonlinear modeling, which collectively improve the robustness of overall predictions. Additionally, the performance of the non-shared AMG is inferior to that of the shared AMG module. This is because the shared AMG module is shared across all layers, facilitating cross-layer and cross-modal instance-level interaction, allowing SCP and VCP to mutually guide each other to capture multi-modal information between semantics and visuals, while also reducing the model space complexity.

\begin{figure*}[ht]
\centering
\includegraphics[width=\textwidth]{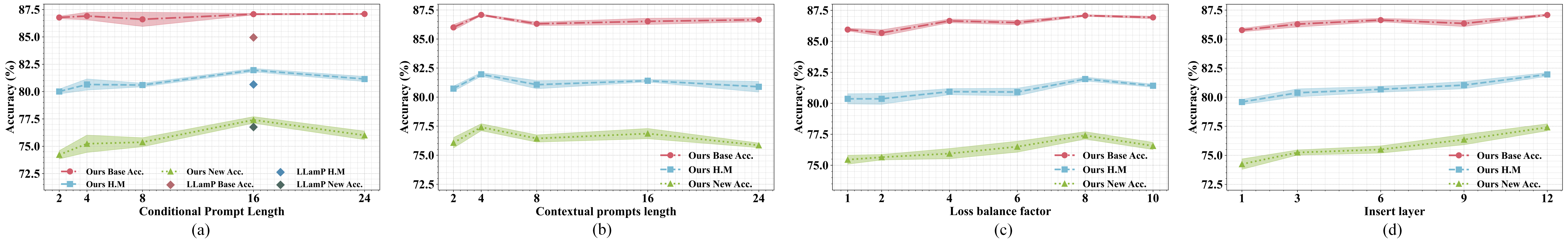} 
\caption{(a) Ablation study of different condition prompt lengths $n$. (b) Ablation study of different contextal prompt lengths $m$. (c) Selection of different loss balance factors $\lambda$. (d) Ablation study of prompt insertion at different layers $l$.  
All results are the average values across 11 different datasets.}
\label{fig:abl-length and hypers}
\end{figure*} 

\begin{table}[ht]
    \centering
    \caption{Ablation study on MPF Mechanism.}
    % \resizebox{\linewidth}{!}{
    \setlength{\tabcolsep}{2.9mm}
    \begin{tabular}{c|cc|cc|c}
        \toprule
        MPF & add & concat & Base & New & HM \\
        \midrule
        \multirow{3}{*}{Ours}
        & \cmark & & 86.96 & 75.61 & 80.89 \\
        & & \cmark & 85.18 & 76.77 & 80.76 \\
        & \cmark & \cmark & \cellcolor{tabhighlight} \bf 87.08 & \cellcolor{tabhighlight} \bf 77.53 & \cellcolor{tabhighlight} \bf 82.03 \\
    \bottomrule
    \end{tabular}
    % } ``concat''
    \label{tab:abl-fusion}
    % \vspace*{-0.28cm}
\end{table}
% \subsection{Effect of MPF Mechanism.}
\noindent\textbf{Effect of MPF Mechanism.} To further validate the impact of the fusion mechanism between conditional prompts and contextual prompts on the final performance, MuGCP examined the commonly used ``add'' operation, ``concat'' operation, and the organic combination of these two operations proposed by us. Specifically, the text embeddings and patch embeddings after the ``add'' operation are represented as follows: $W_{i}^{\textrm{mllms}} = \{P_{\mathcal{T}}^{i}, \cdots, P_{\mathcal{TA}}^{i}\}$ and $E_{i}^{\textrm{mllms}} = \{\cdots, P_{\mathcal{V}}^{i}, P_{\mathcal{VA}}^{i}\}$; and those after the ``concat'' operation are represented as follows: $W_{i}^{\textrm{mllms}} = \{P_{\mathcal{T}}^{i}, \cdots, P_{\mathcal{TD}}^{i}\}$ and $E_{i}^{\textrm{mllms}} = \{\cdots, P_{\mathcal{V}}^{i}, P_{\mathcal{VC}}^{i}\}$. As depicted in Table~\ref{tab:abl-fusion}, the ``add'' operation tends to cause the contextual prompts to overfit to the base classes in the training domain, failing to sufficiently capture useful information from the conditional prompts in the test domain. On the other hand, the ``concat'' operation leverages frozen transformer layers within the encoder to enable complex multi-prompt interactions and progressively learns discriminative features, thereby improving generalization to new classes. Our MPF mechanism integrates and balances these two operations, effectively fusing SCP $P_{\mathcal{TD}}^{i}$ and VCP $P_{\mathcal{VC}}^{i}$ with their corresponding text contextual prompts $P_{\mathcal{T}}^{i}$ and visual contextual prompts $P_{\mathcal{V}}^{i}$. This enhances the learning and modeling of class embeddings and instance-specific knowledge, achieving optimal performance in both base and new classes generalization.

\begin{table}[h]
    \centering
    \caption{Ablation study on more text augmentations. GPT-3 is a text sentence provided by CoPrompt~\cite{roy2023consistency}; GPT-4 comes from LLamP~\cite{zheng2023large}, which directly provides the text embeddings; HT indicates that hand-crafted templates are used for PSRC~\cite{khattak2023self}.}
    \label{tab:abl-text}
    \setlength{\tabcolsep}{0.7mm}
    \begin{tabular}{c|ccccc|cc|c}
        \toprule
        Method & GPT-3 & GPT-4 & $D^{\textrm{llm}}_{K}$ & $D^{\textrm{mllms}}_{K}$ & HT & Base & New & HM \\
        \midrule
         CoPrompt~\cite{roy2023consistency} & \cmark &  &  &  &  & 82.91 & 75.11 & 78.58 \\
         LLamP~\cite{zheng2023large} &  & \cmark &  &  &  & 84.95 & 76.76 & 80.65 \\
         PSRC~\cite{khattak2023self} &  &  &  &  & \cmark & 84.26 & 76.10 & 79.97 \\
         \midrule
         \multirow{6}{*}{Ours} 
         & \cmark &  &  &  &  & 86.97 & 76.75 & 81.54 \\
         &  & \cmark &  &  &  & 86.87 & 76.73 & 81.49 \\
         &  &  & \cmark &  &  & 86.83 & 76.92 & 81.58 \\
         &  &  &  & \cmark &  & \bf 87.20 & 77.04 & 81.81 \\
         &  &  &  &  & \cmark & 86.20 & 76.61 & 81.12 \\
         % \rowcolor{tabhighlight}
         % \midrule
         &  &  & \cmark & \cmark &  & \cellcolor{tabhighlight} 87.08 & \cellcolor{tabhighlight} \bf 77.53 & \cellcolor{tabhighlight} \bf 82.03 \\
    \bottomrule
    \end{tabular}
\end{table}

% \subsection{Effect of More Augmented Text.}
\noindent\textbf{Effect of Augmented Text.} Using augmented text to strengthen consistency constraints between the training model and the pre-trained model has proven to be an effective method to prevent overfitting in prompt learning~\cite{roy2023consistency,zheng2023large,khattak2023self}. Our MuGCP explores two ways to generate augmented texts, $D^{\textrm{llm}}_{K}$ and $D^{\textrm{mllms}}_{K}$, and adopts a strategy of randomly selecting a single sentence embedding for supervision during training, rather than using the averaged embeddings of multiple sentences as in LLamP~\cite{zheng2023large} and PSRC~\cite{khattak2023self}. As shown in Table~\ref{tab:abl-text}, even when MuGCP uses augmented texts from other methods, its performance still surpasses those methods. The fixed average embeddings provided by LLamP and PSRC slightly limit MuGCP's performance on new classes. Compared with CoPrompt~\cite{roy2023consistency}, MuGCP achieves improvements of 4.06\%, 1.64\%, and 2.68\% in base classes accuracy, new classes accuracy, and HM, respectively, when using CoPrompt's augmented texts, highlighting the architectural advantages of our approach. Moreover, we observe that using our designed augmented texts $D^{\textrm{llm}}_{K}$ and $D^{\textrm{mllms}}_{K}$ further enhances MuGCP's performance, indicating that MLLM-based augmented texts provide more sophisticated and high-level semantic supervisory information.

\begin{table*}[t]
    \centering
    % \caption{Comparisons with the latest advancements in MLLMs methods.}
    \caption{Compare with the latest advancements in MLLMs methods under the few-shot learning setting.}
    % \resizebox{\linewidth}{!}{
    \setlength{\tabcolsep}{1.45mm}
    \begin{tabular}{ccccccccc}
        \toprule
        Method & MLLMs & Train samples & Params & ImageNet Train Times & ImageNet & Flowers & Stanford Cars & Caltech101 \\
        \midrule
        VLMClassifier~\cite{zhang2024visually} & BLIP-2-2.7b & zero-shot & / & / & 25.30 & 27.00 & 00.00 & 46.90\\
        VLMClassifier~\cite{zhang2024visually} & BLIP-2-2.7b & all-shots & 105M & 120 hours on a single L40s GPU  & 88.00 & 99.00 & 93.90 & 98.80\\
        Ours & BLIP-2-2.7b & 16-shots & 47.8M & 12 hours on a single RTX3090 GPU & 77.40 & 99.20 & 93.20 & 98.03\\
    \bottomrule
    \end{tabular}
    \label{tab:abl-mllms}
\end{table*}

\noindent\textbf{Effect of Hyperparameters.} As illustrated in Fig.~\ref{fig:abl-length and hypers}, the impact of conditional prompt length $n$, contextual prompt length $m$, loss balance factors $\lambda$, and prompt insertion layers $l$ on the performance of MuGCP is analyzed. Fig.~\ref{fig:abl-length and hypers}(a) demonstrates the effect of conditional prompt length $n$ on MuGCP. While the accuracy on base classes remains consistent as the conditional prompt length increases, the accuracy on new classes improves and peaks at \(n=16\). Fig.~\ref{fig:abl-length and hypers}(b) shows the influence of contextual prompt length $m$. Both base and new classes accuracies reach their peak at \(m=4\), but the generalization performance declines with further increases in length. Fig.~\ref{fig:abl-length and hypers}(c) highlights the sensitivity of MuGCP to the loss balance factors $\lambda$ in the consistency loss. Our results show that increasing $\lambda$ generally improves accuracy, with optimal performance achieved at $\lambda=8$. MuGCP inserts SCP, VCP, and contextual prompts into specific layers of the VLMs, analyzing the effect of inserting prompts at different layers, as shown in Fig.~\ref{fig:abl-length and hypers}(d). The performance improves with deeper insertion layers, with optimal results achieved when conditional prompt learning is applied across layers 1–12. This is because applying such prompt learning in earlier layers ($l<6$) struggles to transfer knowledge effectively to the final text embeddings.

\begin{table}[ht]
    \centering
    \caption{Comparison of different initialization templates.}
    % \resizebox{\linewidth}{!}{
    \setlength{\tabcolsep}{2mm}
    \begin{tabular}{c|ccc}
        \toprule
        initialization templates & Base & New & HM \\
        \midrule
          ``a photo of a $\left \{  \right \}$ '' & 86.41 & 76.81 & 81.33 \\
         % \rowcolor{tabhighlight}
         % \midrule
           ``X X X X $\left \{  \right \}$ ''   & \cellcolor{tabhighlight} \bf 87.08 & \cellcolor{tabhighlight} \bf 77.53 & \cellcolor{tabhighlight} \bf 82.03 \\
    \bottomrule
    \end{tabular}
    % }
    \label{tab:abl-template}
\end{table}
\noindent\textbf{Effect of Different Initialization Templates.} Unlike most methods that use handcrafted templates such as ``a photo of a $\left \{  \right \}$ '' for initialization, we use a randomly initialized template (``X X X X $\left \{  \right \}$ ''). The domain-shared prompts initialized with random templates provide complementary knowledge to the multi-modal conditional prompts, which MuGCP utilizes to enhance domain-shared prompts, leading to better performance.

\noindent\textbf{Comparisons with the MLLMs methods.} We compared MuGCP with VLMClassifier~\cite{zhang2024visually}, which directly utilizes MLLMs for image classification tasks. As shown in Table~\ref{tab:abl-mllms}, directly applying MLLMs for zero-shot image classification yields poor performance. VLMClassifier argues that effective decoding of the internal feature representations of MLLMs for classification requires extensive fine-tuning with a large number of training samples. Consequently, in Table~\ref{tab:abl-mllms}, VLMClassifier uses all-shots training data to fine-tune the Projection layer of BLIP-2 on a single L40S GPU, training for 120 hours to achieve notable performance, demonstrating that key classification information is encoded within the latent space of MLLMs. In comparison, our MuGCP explores a novel approach that efficiently integrates MLLMs and VLMs. For MLLMs, we employ few-shot data to fine-tune and decode their latent key classification information, generating SCP enriched with fine-grained high-level semantic knowledge to enhance the downstream task generalization of VLMs. MuGCP achieves classification performance comparable to MLLMs fine-tuned on all-shots data while requiring significantly less training time (12 hours on a single RTX 3090 GPU) and fewer parameters.

\begin{table}[ht]
    \centering
    \caption{Space and time complexity on the Stanford Cars dataset.}
    \setlength{\tabcolsep}{2mm}
    \begin{tabular}{c|ccccc}
        \toprule
        Methods & Params & Train Times & Base & New & HM \\
        \midrule
          LLamP~\cite{zheng2023large} & 52.8M & 22min & 81.22 & 73.96 & 77.42 \\
           Ours 
           &  47.8M 
           &  22min 
           &  \bf 89.00 
           &  \bf 77.66 
           &  \bf 82.94 \\
    \bottomrule
    \end{tabular}
    \label{tab:abl-complexity}
\end{table}
\noindent\textbf{Space and time complexity.} As shown in Table~\ref{tab:abl-complexity}, we compare the space and time complexity of MuGCP with the state-of-the-art method LLamP on the Stanford Cars dataset. While maintaining similar time complexity, MuGCP demonstrates significantly lower space complexity than LLamP. Furthermore, MuGCP achieves improvements of 7.78\%, 3.7\%, and 5.52\% over LLamP in base classes accuracy, new classes accuracy, and HM, respectively. Although MLLMs leverage few-shot learning to dynamically decode key information within the internal space, providing additional fine-grained high-level semantic knowledge, this leads to an increase in learnable parameters, achieving improved generalization performance at the cost of additional computational resources. In the future, we plan to explore lightweight, training-free methods to integrate MLLM knowledge into VLMs more efficiently.

\begin{figure}[ht]
\centering
\includegraphics[width=0.48\textwidth]{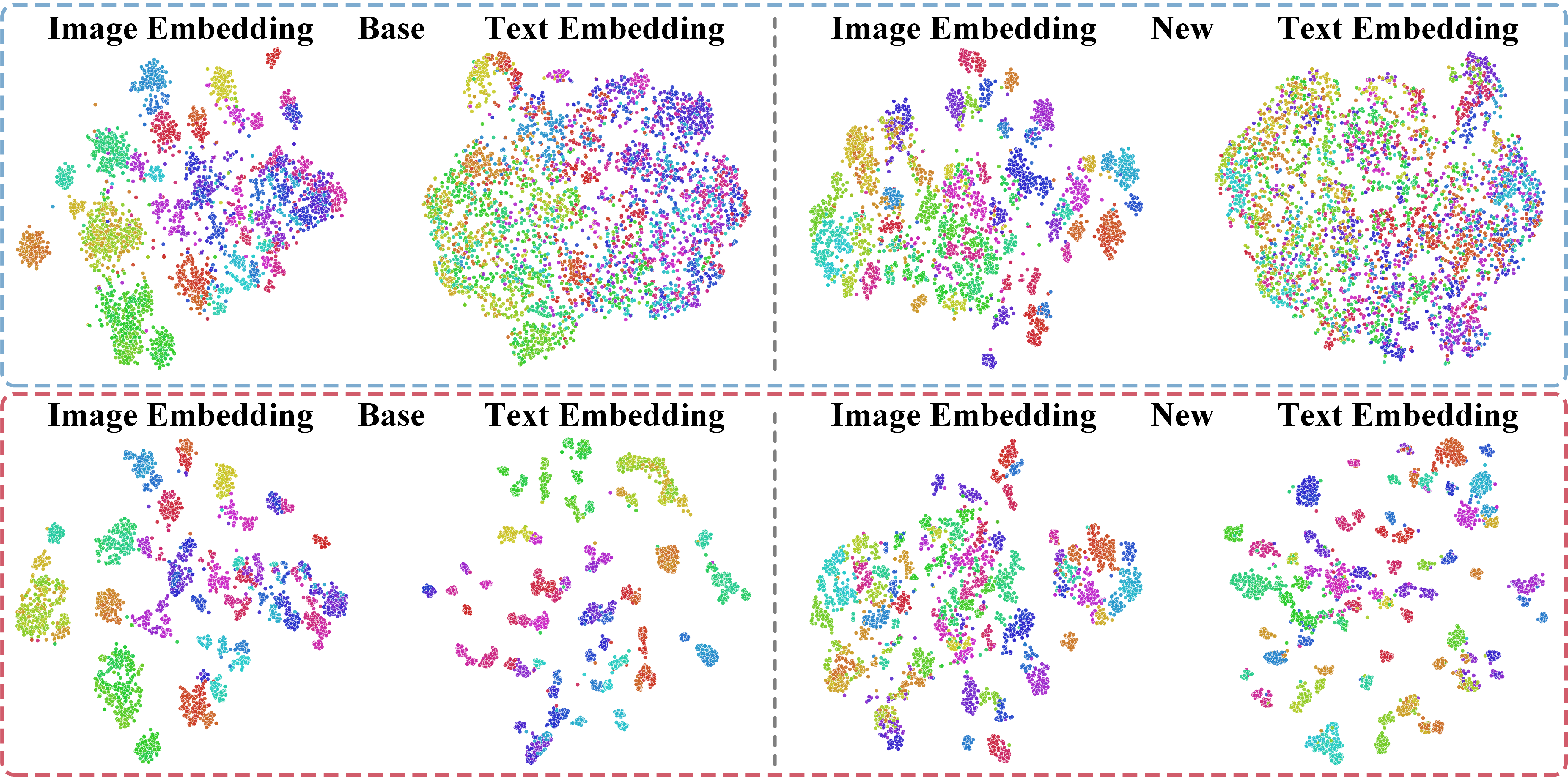} 
\caption{T-SNE plots of image and text embeddings in conditional prompting methods CoCoOp (top) and Our MuGCP (bottom) on the Stanford Cars dataset.} 
\label{fig:visualization}
% \vspace*{-0.28cm}
\end{figure} 

\noindent\textbf{T-SNE Visualization.} To demonstrate the superiority of MuGCP in dynamically adjusting the text and vision representation spaces, we further visualize the image and text embeddings for both base and new classes on the Stanford Cars dataset. As illustrated in Fig.~\ref{fig:visualization}, MuGCP exhibits significant inter-class separability in text embedding for both base and new classes, indicating that our SCP provides rich and fine-grained high-level semantic knowledge. This enables the construction of a discriminative text classifier for prediction. Additionally, our VCP enhances the visual representation space, resulting in more separable image embedding.

\section{Conclusions}
In this paper, we propose MuGCP, which dynamically integrates the encyclopedic knowledge of MLLMs into VLMs for zero/few-shot image classification tasks. Specifically, we innovatively leverage MLLMs as conditional prompt learners to adaptively generate Semantic Conditional Prompts (SCP). Concurrently, we design an Attention Mutual-Guidance (AMG) module to produce Visual Conditional Prompts (VCP) and facilitate cross-layer and cross-modal prompt-level interactions between semantic and visual conditional prompts, dynamically bridging and guiding multi-modal spaces. Furthermore, we introduce a Multi-Prompt Fusion (MPF) mechanism to ensure effective coordination between prompts and embedding information, enhancing the learning and modeling of class embeddings and instance-specific knowledge. Finally, the designed MLLMs-based text augmentation and consistency loss constraint training. Ablation studies and performance comparisons across 14 datasets demonstrate the effectiveness of MuGCP in few-shot learning, base-to-new generalization, domain generalization, and cross-dataset generalization.

\bibliography{tcsvt.bib}
\bibliographystyle{IEEEtran}

\vfill

\newpage
{\appendix[]
Here, we provide additional discussions and detailed results that could not be included in the main paper due to space constraints, including:

\textbf{A.} Detailed experimental results and analysis of Base-to-New generalization for 1/2/4/8/16 shots;

\textbf{B.} Detailed experimental results and analysis of Few-shot Learning for 1/2/4/8/16 shots;

\textbf{C.} Visualization and analysis of T-SNE and Grad-CAM heatmaps for multiple datasets.

\subsection{Detailed Base-to-New Generalization.} 
Table~\ref{tab:1248base-to-new1}, Table~\ref{tab:1248base-to-new2}, Table~\ref{tab:1248base-to-new3}, and Table~\ref{tab:1248base-to-new4} respectively present the numerical results and standard deviation analysis for 1/2/4/8/16-shots in the Base-to-New generalization setting across all datasets. Our method outperforms previous approaches on most datasets. On several datasets, including SUN397, DTD, Aircraft, and Stanford Cars, we significantly surpass previous methods across all-shots settings. Overall, we achieve excellent average performance.

\subsection{Detailed Few-shot Learning.}
Table~\ref{tab:few-shot} present the numerical results and standard deviation analysis across all datasets in the Few-shot Learning setting. Our method outperforms prior methods on most datasets. On several datasets such as Stanford Cars, Aircraft, and DTD, we surpass previous methods by a decent margin across all shot settings. Overall, we achieve a superior average performance.

\subsection{Visualization.}
\noindent\textbf{T-SNE Visualization.}
To demonstrate the superiority of the proposed MuGCP in dynamically and adjusting the text and visual representation spaces for each instance, we further visualize the image and text embeddings for both base and novel classes across multiple datasets. As shown in Figure~\ref{fig:t-sne-sp},
MuGCP exhibits significant inter-class separability in text embeddings for both base and new classes, indicating that our text conditional prompts provide rich and distinctive semantic knowledge. 
This enables the construction of a discriminative text classifier for prediction. Additionally, our visual conditional prompts enhance the visual representation space, resulting in more separable image embeddings that better adapt to CLIP's multimodal space.

\noindent\textbf{Grad-CAM Heatmaps Visualization.}
During testing on new classes, we extract features from the attention blocks of the image encoder to analyze the regions of focus in the model. As depicted in Figure~\ref{fig:cam-sp}, we visualize the Grad-CAM heatmaps across 11 different datasets using various methods. It can be observed that our MuGCP concentrates more on the unique information of the instances themselves, thereby reducing interference from task-irrelevant visual concepts. This is particularly evident in fine-grained datasets such as StanfordCars and Aircraft, where it captures distinctive discriminative information, leading to better generalization.

\begin{table*}[b]
\caption{Ablation Study on MLLMs. \xmark{} refers to ``w/o MLLMs, only initialize $P_{\mathcal{D}}$ with a normal distribution''. $\ddagger$ indicates replacing the image-conditional prompts in CoCoOp with our SCP, while keeping everything consistent with CoCoOp.}
% \fontsize{9pt}{9pt}\selectfont
    \setlength{\tabcolsep}{1.4mm}
    \subfloat{
        \centering
\begin{tabular}{lcccc|ccc|ccc|ccc}
            \toprule 
            \multirow{2}[3]{*}{Method} & \multirow{2}[3]{*}{MLLMs} & \multicolumn{3}{c}{Average} & \multicolumn{3}{c}{ImageNet} & \multicolumn{3}{c}{Caltech101} & \multicolumn{3}{c}{OxfordPets} \\
            \cmidrule(lr){3-5}\cmidrule(lr){6-8}\cmidrule(lr){9-11}\cmidrule(lr){12-14}
            & & Base & New & \multicolumn{1}{c}{HM} & Base & New &\multicolumn{1}{c}{HM} & Base & New & \multicolumn{1}{c}{HM} & Base & New & \multicolumn{1}{c}{HM}\\
            \midrule
            CLIP & /
            & 69.34 & 74.22 & 71.70 
            & 72.43 & 68.14 & 70.22 
            & 96.84 & {94.00} & 95.40 
            & 91.17 & 97.26 & 94.12\\
            \midrule
            CoCoOp & /
            & 80.47 & 71.69 & 75.83 
            & \bf75.98 & {70.43} & {73.10} 
            & 97.96 & 93.81 & {95.84} 
            & 95.20 &\bf 97.69 & {96.43}\\
            \multirow{2}{*}{$\text{CoCoOp}^{\ddagger}$} 
            & \multirow{2}{*}{BLIP-2}
             & \bf80.96 & 73.40 &  77.00
             & 75.73 & 70.37 &  72.95
             & \bf98.37 & \bf94.63 &  \bf96.43
             & \bf95.35 & 97.65 &  \bf96.49
             \\ &
            &  ($\pm0.10 $) &  ($\pm0.75 $) &  ($\pm0.46 $) 
            &  ($\pm0.09 $) &  ($\pm0.17 $) &  ($\pm0.11 $)
            &  ($\pm0.19 $) &  ($\pm0.60 $) &  ($\pm0.40 $)
            &  ($\pm0.15 $) &  ($\pm0.25 $) &  ($\pm0.16 $)
            \\
            \multirow{2}{*}{$\text{CoCoOp}^{\ddagger}$} 
            & \multirow{2}{*}{MiniGPT-4}
             & 80.64 & \bf73.73 &  \bf77.03
             & 75.97 & \bf70.63 &  \bf73.20
             & 98.27 & 94.43 &  96.31
             & 95.33 & 97.50 &  96.40\\ &
            &  ($\pm0.57 $) &  ($\pm1.01 $) &  ($\pm0.45 $)
            &  ($\pm0.12 $) &  ($\pm0.21 $) &  ($\pm0.17 $)
            &  ($\pm0.29 $) &  ($\pm0.69 $) &  ($\pm0.22 $)
            &  ($\pm0.05 $) &  ($\pm0.51 $) &  ($\pm0.27 $) \\
\bottomrule
        \end{tabular}
    }
    \\
    \vspace*{-0.4cm}
    \subfloat{
        \centering
        \begin{tabular}{lcccc|ccc|ccc|ccc}
            \toprule 
            \multirow{2}[3]{*}{Method} & \multirow{2}[3]{*}{MLLMs} & \multicolumn{3}{c}{Stanford Cars} & \multicolumn{3}{c}{Flowers} & \multicolumn{3}{c}{Food101} & \multicolumn{3}{c}{Aircraft} \\
            \cmidrule(lr){3-5}\cmidrule(lr){6-8}\cmidrule(lr){9-11}\cmidrule(lr){12-14}
            & & Base & New & \multicolumn{1}{c}{HM} & Base & New &\multicolumn{1}{c}{HM} & Base & New & \multicolumn{1}{c}{HM} & Base & New & \multicolumn{1}{c}{HM}\\
            \midrule
            CLIP & /
            & 63.37 & 74.89 & 68.65 
            & 72.08 & 77.80 & 74.83 
            & 90.10 & 91.22 & 90.66 
            & 27.19 & 36.29 & 31.09  \\
            \midrule
            CoCoOp  & /
            & 70.49 & 73.59 & 72.01 
            & \bf94.87 & 71.75 & 81.71 
            & \bf\bf90.70 & 91.29 & 90.99
            & 33.41 & 23.71 & 27.74 \\
            \multirow{2}{*}{$\text{CoCoOp}^{\ddagger}$} 
            & \multirow{2}{*}{BLIP-2}
             & 70.70 & \bf74.10 & 72.36
             & 94.17 & \bf73.90 & \bf82.81 
             & 90.40 & \bf91.75 & \bf91.07 
             & \bf35.55 & 32.90 & \bf34.17 \\ &
            &  ($\pm0.30 $) &  ($\pm0.20 $) &  ($\pm0.21 $)
            &  ($\pm0.24 $) &  ($\pm1.91 $) &  ($\pm1.14 $)
            &  ($\pm0.00 $) &  ($\pm0.05 $) &  ($\pm0.02 $)
            &  ($\pm1.15 $) &  ($\pm0.00 $) &  ($\pm0.43 $) \\
            \multirow{2}{*}{$\text{CoCoOp}^{\ddagger}$} 
            & \multirow{2}{*}{MiniGPT-4}
             & \bf72.43 & 72.80 & \bf72.61
             & 93.80 & 73.50 & 82.42
             & 90.60 & 91.53 & 91.06
             & 34.33 & \bf33.00 & 33.50 \\ &
            &  ($\pm1.76 $) &  ($\pm1.65 $) &  ($\pm0.04 $)
            &  ($\pm0.51 $) &  ($\pm1.56 $) &  ($\pm1.05 $)
            &  ($\pm0.08 $) &  ($\pm0.46 $) &  ($\pm0.19 $)
            &  ($\pm2.45 $) &  ($\pm1.40 $) &  ($\pm1.02 $) \\
\bottomrule
        \end{tabular}
    }
    \\
    \vspace*{-0.4cm}
   \subfloat{
        \centering
        \begin{tabular}{lcccc|ccc|ccc|ccc}
            \toprule 
            \multirow{2}[3]{*}{Method} & \multirow{2}[3]{*}{MLLMs} & \multicolumn{3}{c}{SUN397} & \multicolumn{3}{c}{DTD} & \multicolumn{3}{c}{EuroSAT} & \multicolumn{3}{c}{UCF101} \\
            \cmidrule(lr){3-5}\cmidrule(lr){6-8}\cmidrule(lr){9-11}\cmidrule(lr){12-14}
            & & Base & New & \multicolumn{1}{c}{HM} & Base & New &\multicolumn{1}{c}{HM} & Base & New & \multicolumn{1}{c}{HM} & Base & New & \multicolumn{1}{c}{HM}\\
            \midrule
            CLIP & /
            & 69.36 & 75.35 & 72.23 
            & 53.24 & 59.90 & 56.37 
            & 56.48 & 64.05 & 60.03 
            & 70.53 & 77.50 & 73.85 \\
            \midrule
            CoCoOp & /
            & 79.74 & 76.86 & \bf78.27 
            & 77.01 & 56.00 & 64.85 
            & 87.49 & 60.04 & 71.21
            & 82.33 & 73.45 & 77.64 \\ 
            \multirow{2}{*}{$\text{CoCoOp}^{\ddagger}$} 
            & \multirow{2}{*}{BLIP-2}
             & \bf79.80 & 76.40 & 78.06
             & \bf78.93 & 54.77 & 64.67
             & 88.67 & \bf65.40 & \bf75.28 
             & \bf82.90 & 75.57 & \bf79.07 \\ &
            &  ($\pm0.22 $) &  ($\pm0.29 $) &  ($\pm0.19 $)
            &  ($\pm2.13 $) &  ($\pm0.95 $) &  ($\pm0.11 $)
            &  ($\pm2.56 $) &  ($\pm3.77 $) &  ($\pm2.98 $)
            &  ($\pm0.78 $) &  ($\pm2.49 $) &  ($\pm1.69 $) \\
            \multirow{2}{*}{$\text{CoCoOp}^{\ddagger}$} 
            & \multirow{2}{*}{MiniGPT-4}
             & 78.90 & \bf77.50 & 78.19
             & 76.93 & \bf56.70 &  \bf65.28
             & \bf89.15 & 62.60 &  73.55
             & 81.70 & \bf75.73 &  78.60\\ &
            &  ($\pm0.24 $) &  ($\pm0.94 $) &  ($\pm0.08 $)
            &  ($\pm1.28 $) &  ($\pm1.57 $) &  ($\pm0.72 $)
            &  ($\pm0.53 $) &  ($\pm5.96 $) &  ($\pm4.23 $)
            &  ($\pm1.24 $) &  ($\pm2.70 $) &  ($\pm1.41 $) \\
\bottomrule
        \end{tabular}
    }
    \label{table:comparision_with_cocoop}
\end{table*}

\begin{table*}[p]
\caption{For 1/2/4/8/16-shots Base-to-New generalization on \textbf{Average, ImageNet, Caltech101} datasets.}
    \setlength{\tabcolsep}{0.6mm}
    \subfloat{
    \centering
    \begin{tabular}
    {lccc|ccc|ccc|ccc|ccc|ccc|ccc}
    \toprule
    & \multicolumn{3}{c}{1-shots} & \multicolumn{3}{c}{2-shots} & \multicolumn{3}{c}{4-shots} & \multicolumn{3}{c}{8-shots}& \multicolumn{3}{c}{16-shots} \\
    \textbf{Average} & Base & New & HM & Base & New & HM & Base & New & HM & Base & New & HM & Base & New & HM \\
    \cmidrule(r){1-1}
    \cmidrule(r){2-4}
    \cmidrule(r){5-7}
    \cmidrule(r){8-10}
    \cmidrule(r){11-13}
    \cmidrule(r){14-16}
    CLIP		
        &  69.34 &  74.22 &  71.70
        &  69.34 &  74.22 &  71.70
        &  69.34 &  74.22 &  71.70
        &  69.34 &  74.22 &  71.70
        &  69.34 &  74.22 &  71.70 \\  
    \cmidrule(r){1-1}
    \cmidrule(r){2-4}
    \cmidrule(r){5-7}
    \cmidrule(r){8-10}
    \cmidrule(r){11-13}
    \cmidrule(r){14-16}
    CoOP*
        & 70.71  & 64.98  & 67.73
        & 74.52  & 63.09  & 68.33
        & 77.52  & 64.07  & 70.15
        & 80.32  & 64.18  & 71.35
        & {82.69} & 63.22 & 71.66  \\
    CoCoOp*
        &  72.07 & 71.80  & 71.93
        &  73.93 & 71.94  & 72.92
        &  76.64 & 72.39  & 74.46
        &  78.56 & 72.51  & 75.41
        & 80.47 & 71.69 & 75.83  \\
        
    MaPLe*
        &  71.76 &  71.78 &  71.77
        &  74.84 &  73.84 &  74.33
        &  77.20 &  73.16 &  75.13
        &  79.98 &  73.45 &  76.57
        & 82.28 &  75.14 &  78.55  \\

    KgCoOp*
        &  75.48 & 73.16  & 74.30
        &  76.89 & 73.67  & 75.25
        &  77.99 & 73.74  & 75.80
        &  79.42 & 73.54  & 76.37
        & 80.73 & 73.60 & 77.00  \\
        
    PSRC*
        &  73.93 &  71.38 & 72.63
        &  77.95 &  73.07 & 75.43
        &  80.43 &  75.47 & 77.87
        &  82.12 &  74.66 & 78.21
        & 84.26 & 76.10 & 79.97 \\
\bottomrule
    \end{tabular}
    }
    \\
     \vspace*{-0.4cm}
    \subfloat{
    \centering
    \begin{tabular}
    {lccc|ccc|ccc|ccc|ccc|ccc|ccc}
    \toprule
    & \multicolumn{3}{c}{1-shots} & \multicolumn{3}{c}{2-shots} & \multicolumn{3}{c}{4-shots} & \multicolumn{3}{c}{8-shots}& \multicolumn{3}{c}{16-shots} \\
    \textbf{ImageNet} & Base & New & HM & Base & New & HM & Base & New & HM & Base & New & HM & Base & New & HM \\
    \cmidrule(r){1-1}
    \cmidrule(r){2-4}
    \cmidrule(r){5-7}
    \cmidrule(r){8-10}
    \cmidrule(r){11-13}
    \cmidrule(r){14-16}
    CLIP		
        &  72.43 &  68.14 &  70.22
        &  72.43 &  68.14 &  70.22
        &  72.43 &  68.14 &  70.22
        &  72.43 &  68.14 &  70.22
        &  72.43 &  68.14 &  70.22 \\      
    \cmidrule(r){1-1}
    \cmidrule(r){2-4}
    \cmidrule(r){5-7}
    \cmidrule(r){8-10}
    \cmidrule(r){11-13}
    \cmidrule(r){14-16}
    CoOP*
        & 65.40  &  54.20 & 59.28
        & 70.03  &  58.13 & 63.53
        & 72.50  &  60.43 & 65.92
        & 74.83  &  63.07 & 68.45
        & 76.47  & 67.88  & 71.92  \\
    CoCoOP*
        & 74.53  & 69.03  & 71.67
        & 74.77  & 69.33  & 71.95
        & 75.47  & 69.80  & 72.52
        & 75.53  & 70.17  & 72.75
        & 75.98  & 70.43  & 73.10  \\
    MaPLe*
        & 74.77  & 70.23  & 72.43 
        & 75.17  & 70.33  & 72.67
        & 75.67  & 70.77  & 73.14
        & 76.13  & 70.87  & 73.41
        & 76.66  & 70.54  & 73.47  \\
    KgCoOP*
        & 74.10  & 68.70  & 71.29
        & 74.73  & 68.87  & 71.68
        & 75.30  & 69.13  & 72.08
        & 75.47  & 69.63  & 72.43
        & 75.83  & 69.96  & 72.78  \\
    PSRC*
        & 73.53  & 68.43  & 70.89
        & 75.07  & 69.20  & 72.02
        & 76.20  & 69.73  & 72.82
        & 77.03  & 70.03  & 73.36
        & 77.60  & 70.73  & 74.01 \\
\bottomrule
    \end{tabular}
    % }
    % \end{subtable}
    }
    \\
     \vspace*{-0.4cm}
    \subfloat{
    \centering
    \begin{tabular}
    {lccc|ccc|ccc|ccc|ccc|ccc|ccc}
    \toprule
    & \multicolumn{3}{c}{1-shots} & \multicolumn{3}{c}{2-shots} & \multicolumn{3}{c}{4-shots} & \multicolumn{3}{c}{8-shots}& \multicolumn{3}{c}{16-shots} \\
    \textbf{Caltech101} & Base & New & HM & Base & New & HM & Base & New & HM & Base & New & HM & Base & New & HM \\
    \cmidrule(r){1-1}
    \cmidrule(r){2-4}
    \cmidrule(r){5-7}
    \cmidrule(r){8-10}
    \cmidrule(r){11-13}
    \cmidrule(r){14-16}
    CLIP		
        &  96.84 &  94.00 &  95.40
        &  96.84 &  94.00 &  95.40
        &  96.84 &  94.00 &  95.40
        &  96.84 &  94.00 &  95.40
        &  96.84 &  94.00 &  95.40 \\      
    \cmidrule(r){1-1}
    \cmidrule(r){2-4}
    \cmidrule(r){5-7}
    \cmidrule(r){8-10}
    \cmidrule(r){11-13}
    \cmidrule(r){14-16}
    CoOP*
        &  96.97 &  93.77 & 95.34
        &  96.97 &  87.07 & 91.75
        &  97.27 &  88.03 & 92.42
        &  97.73 &  89.97 & 93.69
        & {98.00} & 89.81 & 93.73  \\
    CoCoOP*
        &  97.80 & 93.63  & 95.67
        &  97.93 & 94.50  & 96.09
        &  98.03 & 94.30  & 96.13
        &  97.97 & 93.87  & 95.88
        & 97.96 & 93.81 & {95.84}  \\
    MaPLe*
        &  97.73 & \bf95.33  & 96.52 
        &  97.73 & 94.50  & 96.09
        &  97.80 & 94.50  & 96.12
        &  98.13 & 95.20  & 96.64
        & 97.74 & {94.36} &  {96.02}  \\
    KgCoOP*
        &  98.10 & 94.57  & 96.30
        &  98.10 & 94.80  & 96.42
        &  98.13 & 94.57  & 96.32
        &  98.13 & 94.23  & 96.14
        & 97.72  & 94.39  & 96.03 \\
    PSRC*
        &  98.13 & 94.53  & 96.30
        &  98.27 & 94.70  & 96.45
        &  98.23 & 94.03  & 96.08 
        &  98.27 & 94.13  & 96.16
        & 98.10 & 94.03 & 96.02  \\
\bottomrule
    \end{tabular}
    % }
    % \end{subtable}
    }
    \label{tab:1248base-to-new1}
\end{table*}

\begin{table*}[p]
    \setlength{\tabcolsep}{0.6mm}
     \caption{For 1/2/4/8/16-shots Base-to-New generalization on \textbf{OxfordPets, Stanford Cars, Flowers} datasets.}
    \subfloat{
    \centering
    \begin{tabular}
    {lccc|ccc|ccc|ccc|ccc|ccc|ccc}
    \toprule
    & \multicolumn{3}{c}{1-shots} & \multicolumn{3}{c}{2-shots} & \multicolumn{3}{c}{4-shots} & \multicolumn{3}{c}{8-shots}& \multicolumn{3}{c}{16-shots} \\
    \textbf{OxfordPets} & Base & New & HM & Base & New & HM & Base & New & HM & Base & New & HM & Base & New & HM \\
    \cmidrule(r){1-1}
    \cmidrule(r){2-4}
    \cmidrule(r){5-7}
    \cmidrule(r){8-10}
    \cmidrule(r){11-13}
    \cmidrule(r){14-16}
    CLIP			
        &  91.17 &  97.26 &  94.12
        &  91.17 &  97.26 &  94.12
        &  91.17 &  97.26 &  94.12
        &  91.17 &  97.26 &  94.12
        &  91.17 &  97.26 &  94.12 \\      
    \cmidrule(r){1-1}
    \cmidrule(r){2-4}
    \cmidrule(r){5-7}
    \cmidrule(r){8-10}
    \cmidrule(r){11-13}
    \cmidrule(r){14-16}
    CoOP*
        &  87.90 &  96.43 & 91.97
        &  89.77 &  96.03 & 92.79
        &  91.40 &  95.70 & 93.50
        &  93.17 &  94.40 & 93.78
        & 93.67 & 95.29 & 94.47 \\
    CoCoOP*
        &  94.00 &  97.43 & 95.68
        &  94.27 &  97.37 & 95.79
        &  95.10 &  \bf97.70 & \bf96.38
        &  95.53 &  \bf97.73 & \bf96.62
        & {95.20} & {97.69} & {96.43}  \\
    MaPLe*
        &  91.97 &  96.97 & 94.40 
        &  93.53 &  97.10 & 95.28
        &  94.67 &  97.30 & 95.97
        &  95.53 &  97.27 & 96.39
        &  {95.43} & {97.76} &  {96.58}  \\
    KgCoOP*
        &  93.57 &  97.33 & 95.41
        &  93.97 &  97.50 & 95.70
        &  94.73 &  97.73 & 96.21
        &  94.93 &  97.63 & 96.26
        & 94.65 & 97.76 & 96.18  \\
    PSRC*
        &  93.73 &  97.37 & 95.52
        &  \bf94.67 &  \bf97.63 & \bf96.13
        &  95.33 &  97.27 & 96.29
        &  95.37 &  95.97 & 95.67
        & 95.33 & 97.30 & 96.30 \\
\bottomrule
    \end{tabular}
    }
     \\
     \vspace*{-0.4cm}
    \subfloat{
    \centering
    \begin{tabular}
    {lccc|ccc|ccc|ccc|ccc|ccc|ccc}
    \toprule
    & \multicolumn{3}{c}{1-shots} & \multicolumn{3}{c}{2-shots} & \multicolumn{3}{c}{4-shots} & \multicolumn{3}{c}{8-shots}& \multicolumn{3}{c}{16-shots} \\
    \textbf{Cars} & Base & New & HM & Base & New & HM & Base & New & HM & Base & New & HM & Base & New & HM \\
    \cmidrule(r){1-1}
    \cmidrule(r){2-4}
    \cmidrule(r){5-7}
    \cmidrule(r){8-10}
    \cmidrule(r){11-13}
    \cmidrule(r){14-16}
    CLIP			
        &  63.37 &  74.89 &  68.65
        &  63.37 &  74.89 &  68.65
        &  63.37 &  74.89 &  68.65
        &  63.37 &  74.89 &  68.65
        &  63.37 &  74.89 &  68.65 \\      
    \cmidrule(r){1-1}
    \cmidrule(r){2-4}
    \cmidrule(r){5-7}
    \cmidrule(r){8-10}
    \cmidrule(r){11-13}
    \cmidrule(r){14-16}
    CoOP*
        &  58.87 &  62.93 & 60.83
        &  65.17 &  62.23 & 63.67
        &  71.17 &  62.30 & 66.44
        &  74.60 &  61.43 & 67.38
        & {78.12} & 60.40 & 68.13 \\
    CoCoOP*
        &  65.77 &  74.13 & 69.70
        &  67.10 &  75.03 & 70.84
        &  67.50 &  74.03 & 70.61
        &  69.83 &  73.27 & 71.51
        & 70.49 & 73.59 & {72.01} \\
    MaPLe*
        &  64.83 &  73.50 & 68.89 
        &  67.10 &  74.77 & 70.73
        &  68.43 &  74.27 & 71.23
        &  69.93 &  75.00 & 72.38
        & 72.94 & 74.00 &  {73.47} \\
    KgCoOP*
        &  66.30 &  75.47 & 70.59
        &  66.03 &  75.97 & 70.65
        &  68.37 &  75.30 & 71.67
        &  70.40 &  75.27 & 72.75
        & 71.76  & {75.04}  & 73.36 \\
    PSRC*
        &  67.83 &  74.03 & 70.79
        &  69.90 &  74.27 & 72.02
        &  72.43 &  75.10 & 73.74
        &  75.63 &  74.73 & 75.18
        & {78.27} & 74.97 & {76.58} \\
\bottomrule
    \end{tabular}
    % }
    }
     \\
     \vspace*{-0.4cm}
    \subfloat{
    \centering
    \begin{tabular}
    {lccc|ccc|ccc|ccc|ccc|ccc|ccc}
    \toprule
    & \multicolumn{3}{c}{1-shots} & \multicolumn{3}{c}{2-shots} & \multicolumn{3}{c}{4-shots} & \multicolumn{3}{c}{8-shots}& \multicolumn{3}{c}{16-shots} \\
    \textbf{Flowers} & Base & New & HM & Base & New & HM & Base & New & HM & Base & New & HM & Base & New & HM \\
    \cmidrule(r){1-1}
    \cmidrule(r){2-4}
    \cmidrule(r){5-7}
    \cmidrule(r){8-10}
    \cmidrule(r){11-13}
    \cmidrule(r){14-16}
    CLIP		
        &  72.08 &  77.80 &  74.83
        &  72.08 &  77.80 &  74.83
        &  72.08 &  77.80 &  74.83
        &  72.08 &  77.80 &  74.83
        &  72.08 &  77.80 &  74.83 \\      
    \cmidrule(r){1-1}
    \cmidrule(r){2-4}
    \cmidrule(r){5-7}
    \cmidrule(r){8-10}
    \cmidrule(r){11-13}
    \cmidrule(r){14-16}
    CoOP*
        &  86.33 &  69.83 & 77.21
        &  90.60 &  68.43 & 77.97
        &  92.43 &  66.77 & 77.53
        &  96.27 &  64.93 & 77.55
        & {97.60} & 59.67 & 74.06 \\
    CoCoOP*
        &  75.87 &  \bf76.83 & 76.35
        &  80.97 & 75.40  & 78.09
        &  86.97 & 72.13  & 78.86
        &  91.40 & 72.23  & 80.69
        & 94.87 & 71.75 & {81.71} \\
    MaPLe*
        &  75.90 &  74.70 & 75.30 
        &  82.83 &  76.20 & 79.38
        &  88.80 &  73.87 & 80.65
        &  93.20 &  74.63 & 82.89
        & 95.92 & 72.46 &  82.56 \\
    KgCoOP*
        & 83.17  &  75.40 & 79.09
        & 85.67  &  75.13 & 80.05
        & 85.40  &  74.90 & 79.81
        & 91.40  &  75.07 & 82.43
        & 95.00  & 74.73  & 83.65 \\
    PSRC*
        & 86.67 & 75.83  & 80.89
        & 90.70  & 74.07  & 81.55
        & 93.03  & 76.37  & 83.88
        & 96.53  & 77.37  & 85.89
        & {98.07} & 76.50 & 85.95 \\
\bottomrule
    \end{tabular}
    % }
    }
    \label{tab:1248base-to-new2}
\end{table*}

\begin{table*}[p]
    \caption{For 1/2/4/8/16-shots Base-to-New generalization on \textbf{Food101, Aircraft, SUN397} datasets.}
    \setlength{\tabcolsep}{0.6mm}
    \subfloat{
    \centering
    \begin{tabular}
    {lccc|ccc|ccc|ccc|ccc|ccc|ccc}
    \toprule
    & \multicolumn{3}{c}{1-shots} & \multicolumn{3}{c}{2-shots} & \multicolumn{3}{c}{4-shots} & \multicolumn{3}{c}{8-shots}& \multicolumn{3}{c}{16-shots} \\
    \textbf{Food101} & Base & New & HM & Base & New & HM & Base & New & HM & Base & New & HM & Base & New & HM \\
    \cmidrule(r){1-1}
    \cmidrule(r){2-4}
    \cmidrule(r){5-7}
    \cmidrule(r){8-10}
    \cmidrule(r){11-13}
    \cmidrule(r){14-16}
    CLIP		
        &  90.10 &  91.22 &  90.66
        &  90.10 &  91.22 &  90.66
        &  90.10 &  91.22 &  90.66
        &  90.10 &  91.22 &  90.66
        &  90.10 &  91.22 &  90.66 \\      
    \cmidrule(r){1-1}
    \cmidrule(r){2-4}
    \cmidrule(r){5-7}
    \cmidrule(r){8-10}
    \cmidrule(r){11-13}
    \cmidrule(r){14-16}
    CoOP*
        &  83.50 & 86.37  & 84.91
        &  83.23 & 82.57  & 82.89
        &  84.93 & 84.17  & 84.55
        &  86.43 & 81.73  & 84.01
        & 88.33 & 82.26 & 85.19 \\
    CoCoOP*
        & 88.93  & 90.27  & 89.59
        & 89.20  & 90.00  & 89.59
        & 89.97  & 90.97  & 90.47
        & 89.97  & 91.10  & 90.53
        & {90.70} & {91.29} & {90.99} \\
    MaPLe*
        & 88.67  &  90.90 & 89.77 
        & 88.03  &  89.47 & 88.74
        & 89.53  &  91.10 & 90.31
        & 89.63  &  91.23 & 90.42
        &  {90.71} &  \bf 92.05 &  91.38 \\
    KgCoOP*
        & \bf89.87  & \bf91.47  & \bf90.66
        &  90.00 &  91.33 & 90.66
        &  \bf90.30 & 91.47  & \bf90.88
        &  \bf90.30 & \bf91.57  & \bf90.93
        & 90.50 &  91.70 & 91.09 \\
    PSRC*
        & 88.13  &  91.00 & 89.54
        & 88.73  & 90.97  & 89.84
        & 89.70  & 91.13  & 90.41
        & 90.00  & 91.53  & 90.76
        & 90.67 & 91.53 & 91.10 \\
\bottomrule
    \end{tabular}
    % }
    }
     \\
     \vspace*{-0.4cm}
    \subfloat{
    \centering
    \begin{tabular}
    {lccc|ccc|ccc|ccc|ccc|ccc|ccc}
    \toprule
    & \multicolumn{3}{c}{1-shots} & \multicolumn{3}{c}{2-shots} & \multicolumn{3}{c}{4-shots} & \multicolumn{3}{c}{8-shots}& \multicolumn{3}{c}{16-shots} \\
    \textbf{Aircraft} & Base & New & HM & Base & New & HM & Base & New & HM & Base & New & HM & Base & New & HM \\
    \cmidrule(r){1-1}
    \cmidrule(r){2-4}
    \cmidrule(r){5-7}
    \cmidrule(r){8-10}
    \cmidrule(r){11-13}
    \cmidrule(r){14-16}
    CLIP			
        &  27.19 &  36.29 &  31.09
        &  27.19 &  36.29 &  31.09
        &  27.19 &  36.29 &  31.09
        &  27.19 &  36.29 &  31.09
        &  27.19 &  36.29 &  31.09\\      
    \cmidrule(r){1-1}
    \cmidrule(r){2-4}
    \cmidrule(r){5-7}
    \cmidrule(r){8-10}
    \cmidrule(r){11-13}
    \cmidrule(r){14-16}
    CoOP*
        &  26.73 &  24.77 & 25.71
        &  29.83 &  19.27 & 23.41
        &  31.90 &  24.87 & 27.95
        &  36.87 &  26.03 & 30.52
        & {40.44} & 22.30 & 28.75 \\
    CoCoOP*
        &  21.90 &  24.37 & 23.07
        &  20.40 &  15.90 & 17.87
        &  32.10 &  34.03 & 33.04
        &  34.63 &  33.77 & 34.19
        & 33.41 & 23.71 & 27.74 \\
    MaPLe*
        &  23.40 & 24.17  & 23.78
        &  31.60 & 35.10  & 33.26
        &  32.40 & 33.53  & 32.96
        &  34.57 & 33.50  & 34.03
        & 37.44 & 35.61 & {36.50}\\
    KgCoOP*
        &  29.13 & 31.43  & 30.24
        &  31.50 & 32.97  & 32.22
        &  32.40 & 33.40  & 32.89
        &  34.77 & 33.30  & 34.02
        & 36.21 & 33.55 & 34.83 \\
    PSRC*
        &  11.00 & 7.00  & 8.56
        &  29.83 & 25.73  & 27.63
        &  35.90 & 36.03  & 35.96
        &  36.70 & 25.57  & 30.14
        & {42.73} & 37.87 & {40.15} \\
\bottomrule
    \end{tabular}
    % }
    }
    \\
     \vspace*{-0.4cm}
    \subfloat{
    \centering
    \begin{tabular}
    {lccc|ccc|ccc|ccc|ccc|ccc|ccc}
    \toprule
    & \multicolumn{3}{c}{1-shots} & \multicolumn{3}{c}{2-shots} & \multicolumn{3}{c}{4-shots} & \multicolumn{3}{c}{8-shots}& \multicolumn{3}{c}{16-shots} \\
    \textbf{SUN397} & Base & New & HM & Base & New & HM & Base & New & HM & Base & New & HM & Base & New & HM \\
    \cmidrule(r){1-1}
    \cmidrule(r){2-4}
    \cmidrule(r){5-7}
    \cmidrule(r){8-10}
    \cmidrule(r){11-13}
    \cmidrule(r){14-16}
    CLIP			
        &  69.36 &  75.35 &  72.23
        &  69.36 &  75.35 &  72.23
        &  69.36 &  75.35 &  72.23
        &  69.36 &  75.35 &  72.23
        &  69.36 &  75.35 &  72.23 \\      
    \cmidrule(r){1-1}
    \cmidrule(r){2-4}
    \cmidrule(r){5-7}
    \cmidrule(r){8-10}
    \cmidrule(r){11-13}
    \cmidrule(r){14-16}
    CoOP*
        &  65.90 &  60.03 & 62.83
        &  71.60 &  59.53 & 65.01
        &  76.00 &  62.03 & 68.31
        &  78.17 &  60.30 & 68.08
        & {80.60} & 65.89 & 72.51 \\
    CoCoOP*
        &  73.77 & 76.30  & 75.01
        &  75.80 & 76.20  & 75.99
        &  76.97 & 76.43  & 76.69
        &  78.50 & 76.33  & 77.39
        & 79.74 & {76.86} & {78.27}\\
    MaPLe*
        &  74.37 & 77.17  & 75.74 
        &  76.47 & 78.30  & 77.37
        &  77.73 & 77.73  & 77.73
        &  79.27 & 77.83  & 78.54
        & {80.82} &  {78.70} &  {79.75} \\
    KgCoOP*
        &  74.60 & 76.97  & 75.77
        &  76.63 & 74.93  & 75.77
        &  78.47 & 76.23  & 77.33
        &  79.43 & 76.87  & 78.13
        & 80.29 & 76.53 & 78.36 \\
    PSRC*
        &  75.67 & 76.73  & 76.19
        &  78.00 & 77.87  & 77.93
        &  80.17 & 78.23  & 79.19
        &  81.60 & 78.10  & 79.81
        & {82.67} & 78.47 & 80.52 \\
\bottomrule
    \end{tabular}
    % }
    }
    \label{tab:1248base-to-new3}
\end{table*}

\begin{table*}[p]
    \caption{For 1/2/4/8/16-shots Base-to-New generalization on \textbf{DTD, EuroSAT, UCF101} datasets.}
    \setlength{\tabcolsep}{0.6mm}
    \subfloat{
    \centering
    \begin{tabular}
    {lccc|ccc|ccc|ccc|ccc|ccc|ccc}
    \toprule
    & \multicolumn{3}{c}{1-shots} & \multicolumn{3}{c}{2-shots} & \multicolumn{3}{c}{4-shots} & \multicolumn{3}{c}{8-shots}& \multicolumn{3}{c}{16-shots} \\
    \textbf{DTD} & Base & New & HM & Base & New & HM & Base & New & HM & Base & New & HM & Base & New & HM \\
    \cmidrule(r){1-1}
    \cmidrule(r){2-4}
    \cmidrule(r){5-7}
    \cmidrule(r){8-10}
    \cmidrule(r){11-13}
    \cmidrule(r){14-16}
    CLIP			
        &  53.24 &  59.90 &  56.37
        &  69.36 &  75.35 &  72.23
        &  53.24 &  59.90 &  56.37
        &  53.24 &  59.90 &  56.37
        &  53.24 &  59.90 &  56.37\\      
    \cmidrule(r){1-1}
    \cmidrule(r){2-4}
    \cmidrule(r){5-7}
    \cmidrule(r){8-10}
    \cmidrule(r){11-13}
    \cmidrule(r){14-16}
    CoOP*
        &  62.33 &  52.30 & 56.88
        &  63.43 &  52.07 & 57.19
        &  69.87 &  49.37 & 57.86
        &  77.30 &  45.23 & 57.07
        & {79.44} & 41.18 & 54.24\\
    CoCoOP*
        &  59.70 &  56.87 & 58.25
        &  62.03 &  55.63 & 58.66
        &  68.07 &  55.10 & 60.90
        &  72.17 &  57.80 & 64.19
        & 77.01 & 56.00 & {64.85} \\
    MaPLe*
        &  57.37 &  54.87 & 56.09
        &  65.07 &  52.80 & 58.30
        &  69.33 &  59.47 & 64.02
        &  75.10 &  55.63 & 63.92
        & {80.36} & 59.18 &  {68.16} \\
    KgCoOP*
        &  67.07 &  57.77 & 62.07
        &  67.30 &  59.30 & 63.05
        &  72.43 &  56.13 & 63.25
        &  75.57 &  58.20 & 65.76
        & 77.55 & 54.99 & 64.35 \\
    PSRC*
        &  67.50 &  55.10 & 60.67
        &  69.07 &  57.10 & 62.52
        &  75.50 &  57.43 & 65.24
        &  80.37 &  62.07 & 70.04
        & {83.37} & 62.97 & 71.75 \\
\bottomrule
    \end{tabular}
    % }
    }
     \\
     \vspace*{-0.4cm}
    \subfloat{
    \centering
    \begin{tabular}
    {lccc|ccc|ccc|ccc|ccc|ccc|ccc}
    \toprule
    & \multicolumn{3}{c}{1-shots} & \multicolumn{3}{c}{2-shots} & \multicolumn{3}{c}{4-shots} & \multicolumn{3}{c}{8-shots}& \multicolumn{3}{c}{16-shots} \\
    \textbf{EuroSAT} & Base & New & HM & Base & New & HM & Base & New & HM & Base & New & HM & Base & New & HM \\
    \cmidrule(r){1-1}
    \cmidrule(r){2-4}
    \cmidrule(r){5-7}
    \cmidrule(r){8-10}
    \cmidrule(r){11-13}
    \cmidrule(r){14-16}
     CLIP			
        &  56.48 &  64.05 & 60.02
        &  56.48 &  64.05 & 60.02
        &  56.48 &  64.05 & 60.02
        &  56.48 &  64.05 & 60.02
        &  56.48 &  64.05 & 60.02 \\      
    \cmidrule(r){1-1}
    \cmidrule(r){2-4}
    \cmidrule(r){5-7}
    \cmidrule(r){8-10}
    \cmidrule(r){11-13}
    \cmidrule(r){14-16}
    CoOP*
        & 70.7  & 47.67  & 56.94
        & 82.33  & 45.20  & 58.36
        & 84.53  & 50.00  & 62.83
        & 85.83  & 56.73  & 68.31
        & {92.19} & 54.74 & 68.69 \\
    CoCoOP*
        &  65.47 & 58.17  & 61.60
        &  74.20 &  67.70 & 70.80
        &  73.83 & 58.03  & 64.93
        &  78.20 & 59.27  & 67.43
        & 87.49 & 60.04 & {71.21} \\
    MaPLe*
        & 65.50  & 56.70  &  60.78
        &  68.40 &  66.30 & 67.33
        &  76.37 & 54.80  & 63.81
        & 86.67  &  59.73 & 70.72
        & {94.07} &  {73.23} & {82.35} \\
    KgCoOP*
        & 77.40  & 57.43  & 65.93
        & 82.93  & 60.43  & 69.91
        & 83.43  & 62.23  & 71.29
        & 81.47  & 59.10  & 68.51
        & 85.64 & 64.34 & 73.48 \\
    PSRC*
        & 73.13  & 66.83  & 69.84
        & 82.07  & 67.60  & 74.14
        & 85.20  & \bf77.33  & \bf81.07
        &  86.37 & 73.90  & 79.65
        & 92.90 & 73.90 & 82.32 \\
\bottomrule
    \end{tabular}
    % }
    }
    \\
     \vspace*{-0.4cm}
    \subfloat{
    \centering
    \begin{tabular}
    {lccc|ccc|ccc|ccc|ccc|ccc|ccc}
    \toprule
    & \multicolumn{3}{c}{1-shots} & \multicolumn{3}{c}{2-shots} & \multicolumn{3}{c}{4-shots} & \multicolumn{3}{c}{8-shots}& \multicolumn{3}{c}{16-shots} \\
    \textbf{UCF101} & Base & New & HM & Base & New & HM & Base & New & HM & Base & New & HM & Base & New & HM \\
    \cmidrule(r){1-1}
    \cmidrule(r){2-4}
    \cmidrule(r){5-7}
    \cmidrule(r){8-10}
    \cmidrule(r){11-13}
    \cmidrule(r){14-16}
    CLIP			
        &  70.53 &  77.50 &  73.85
        &  70.53 &  77.50 &  73.85
        &  70.53 &  77.50 &  73.85
        &  70.53 &  77.50 &  73.85
        &  70.53 &  77.50 &  73.85 \\      
    \cmidrule(r){1-1}
    \cmidrule(r){2-4}
    \cmidrule(r){5-7}
    \cmidrule(r){8-10}
    \cmidrule(r){11-13}
    \cmidrule(r){14-16}
    CoOP*
        & 73.20  & 66.50  & 69.69
        &  76.80 &  63.43 & 69.48
        & 80.67  & 61.10  & 69.53
        & 82.30  & 62.20  & 70.85
        & {84.69} & 56.05 & 67.46 \\
    CoCoOP*
        & 75.03  & 72.77  & 73.88
        & 76.63  & 74.27  & 75.43
        & 79.07  & 73.87  & 76.38
        & 80.40  & 72.07  & 76.01
        & 82.33 & 73.45 & {77.64}  \\
    MaPLe*
        & 74.80  & 75.07  &  74.93
        & 77.30  & 77.33  & 77.32
        & 78.50  & 77.50  & 77.99
        & 81.57  & 77.03  & 79.24
        & 83.00 &  {78.66} & {80.77} \\
    KgCoOP*
        & 76.93  & 78.23  & 77.57
        & 79.03  & \bf79.17  & 79.09
        & 78.97  & \bf80.00  & 79.48
        & 81.77  & 78.07  & 79.88
        & 82.89 & 76.67  & 79.65 \\
    PSRC*
        & 77.87  & 78.30  & 78.08
        &  81.17 & 74.67  & 77.78
        & 83.00  & 77.57  & 80.19
        & 85.47  & 77.83  & 81.47
        & {87.10} & 78.80 & 82.74 \\
\bottomrule
    \end{tabular}
    % }
    }
    \label{tab:1248base-to-new4}
\end{table*}

\begin{table*}[p]
\caption{Few shot classification all results and Std with 1/2/4/8/16-shots. $\bm{\triangle}$ represents the gain relative to LLamP.}
% \fontsize{9pt}{9pt}\selectfont
\setlength{\tabcolsep}{0.8mm}
\begin{tabular}{l|c|cccccccccccc}
\toprule
Methods & Shots                                          
& \rotatebox{90}{Average} 
     & \rotatebox{90}{ImageNet} 
     & \rotatebox{90}{Caltech101} 
     & \rotatebox{90}{OxfordPets} 
     & \rotatebox{90}{StanfordCars} 
     & \rotatebox{90}{Flowers} 
     & \rotatebox{90}{Food101} 
     & \rotatebox{90}{Aircraft} 
     & \rotatebox{90}{SUN397} 
     & \rotatebox{90}{DTD} 
     & \rotatebox{90}{EuroSAT} 
     & \rotatebox{90}{UCF101} 

\\ 
% \midrule
\bottomrule
\addlinespace
zero-shot CLIP & & 65.42 & 66.70 & 94.20 & 89.10 & 65.50 & 70.80 & 85.90 & 24.80 & 62.60 & 44.00 & 48.40 & 67.60 \\ 
% \midrule
\bottomrule
\addlinespace
Linear probe  & \multirow{7}{*}{1-shots} & 45.83 & 32.13 & 79.88 & 44.06 & 35.66 & 69.74 & 43.96 & 19.61 & 41.58 & 34.59 & 49.23 & 53.66 \\
CoOp &  & 67.56 & 66.33 & 92.60 & 90.37 & 67.43 & 77.53 & 84.33 & 21.37 & 66.77 & 50.23 & 54.93 & 71.23 \\
CoCoOp  &  & 69.10 & 69.50 & 93.30 & 91.80 & 67.50 & 73.80 & 84.90 & 28.20 & 68.50 & 53.30 & 58.60 & 70.60 \\
MaPLe & & 69.27 & 62.67 & 92.57 & 89.10 & 66.60 & 83.30 & 80.50 & 26.73 & 64.77 & 52.13 & 71.80 & 71.83 \\
PSRC & & 72.32 & 68.13 & 93.67 & 92.00 & 69.40 & 85.93 & 84.87 & 27.67 & 69.67 &  56.23 & 73.13 & 74.80 \\
TGP-T-F & & 72.15 & 69.84 & 94.20 & 88.28 & 70.44 & 85.67 & 84.49 & 30.36 & 69.30 &  54.73 & 72.96 & 73.35 \\
\bottomrule
\end{tabular}
\label{tab:few-shot}
\end{table*}

\begin{figure*}[b]
\centering
\includegraphics[width=1.00\textwidth]{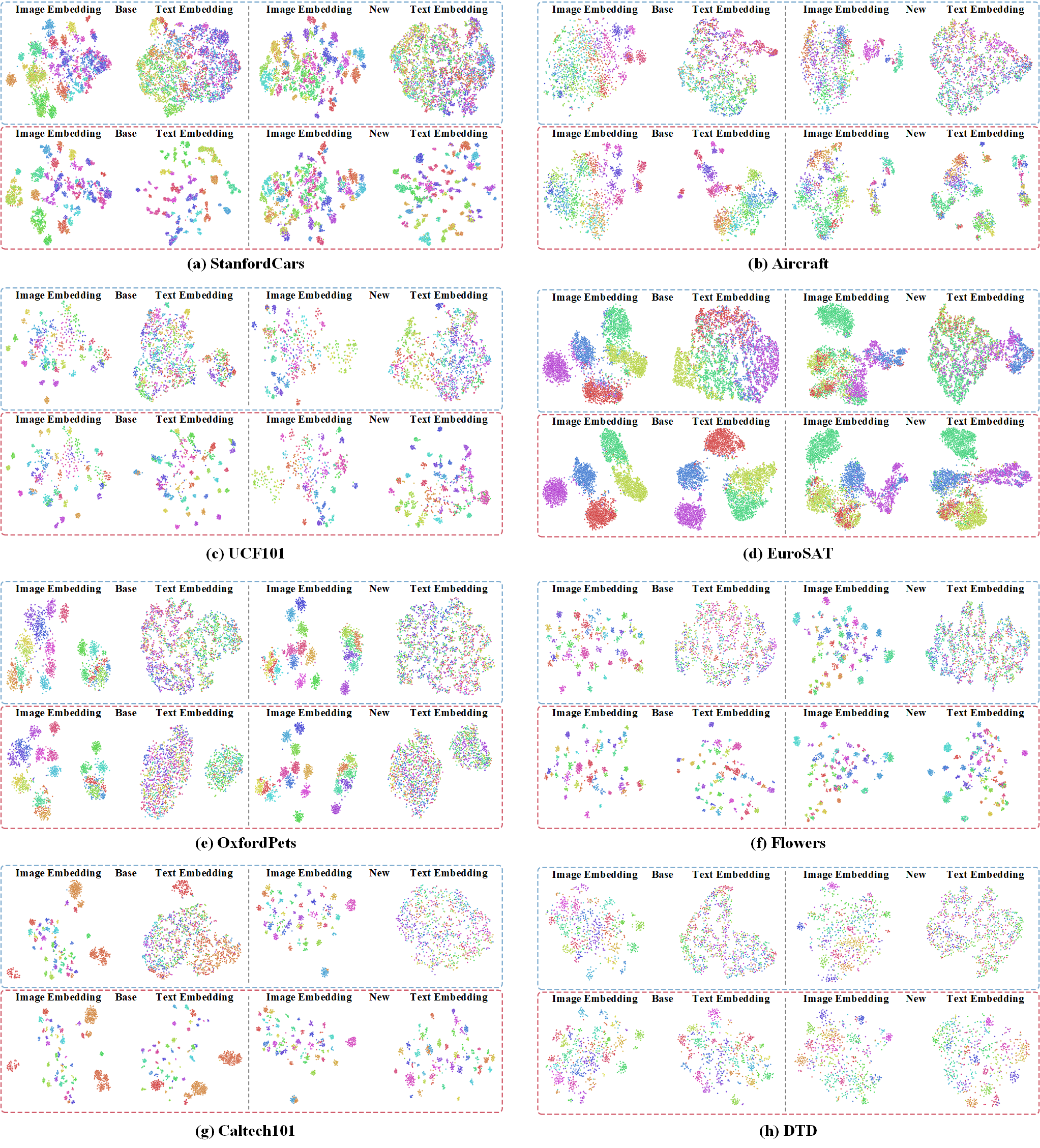}
\caption{T-SNE plots of image embeddings and text embeddings in conditional prompting method CoCoOp (top) and Our MuGCP (bottom) on \textbf{Stanford Cars, Aircraft, UCF101, EuroSAT, OxfordPets, Flowers, Caltech101, DTD} datasets.}
\label{fig:t-sne-sp}
\end{figure*}

\begin{figure*}[b]
\centering
\includegraphics[width=1.00\textwidth]{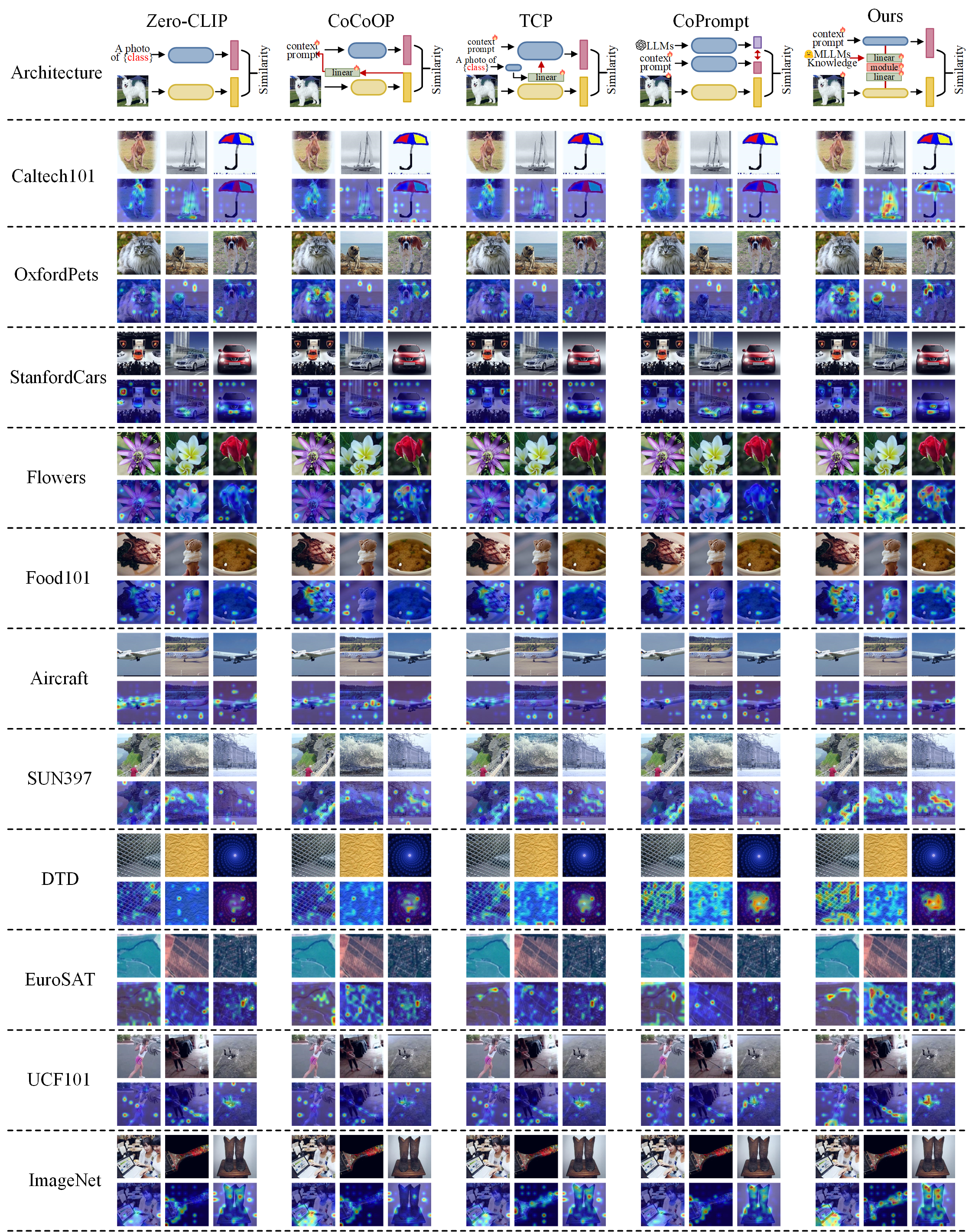}
\caption{Grad-CAM heatmaps for new class generalization on 11 different image recognition datasets for \textbf{CLIP~\cite{radford2021learningCLIP}, CoCoOp~\cite{zhou2022conditionalCoCoOp}, TCP~\cite{yao2024tcp}, CoPrompt~\cite{roy2023consistency}, and our MuGCP}.}
\label{fig:cam-sp}
\end{figure*}

\end{document}